\documentclass{article}

 \usepackage[preprint]{neurips_2026}

\usepackage{float}
\usepackage[utf8]{inputenc} 
\usepackage[T1]{fontenc}    
\usepackage[dvipsnames]{xcolor}
\usepackage{amsmath}
\usepackage{amssymb}
\usepackage{mathtools}
\usepackage{amsthm}
\usepackage{adjustbox}
\usepackage{times}
\usepackage{soul}
\usepackage{url}
\usepackage{xcolor}
\usepackage[most]{tcolorbox}
\usepackage{booktabs}
\usepackage{mathtools}
\usepackage{amsfonts}
\usepackage{amssymb}
\usepackage{makecell}
\usepackage{hyperref} 
\usepackage{float}
\usepackage{algorithmic}
\usepackage{graphicx}
\usepackage{multirow}
\usepackage{multicol}
\usepackage{algorithm2e}
\usepackage{graphicx}
\usepackage{subcaption}
\usepackage[table]{xcolor}
\usepackage{booktabs}
\usepackage{multirow}
\usepackage{colortbl}
\usepackage{xcolor}
\usepackage{makecell}
\usepackage{amssymb}
\usepackage{pifont}
\definecolor{xred}{HTML}{C62828}
\definecolor{checkgreen}{HTML}{2E7D32}
\newcommand{\xmark}{\textcolor{xred}{\ding{55}}}
\newcommand{\cmark}{\textcolor{checkgreen}{\ding{51}}}
\definecolor{oursgreen}{HTML}{E8F5E9}
\definecolor{ourstextgreen}{HTML}{2E7D32}
\definecolor{headerblue}{HTML}{E3F2FD}
\definecolor{capgray}{HTML}{F5F5F5}
\title{Beyond Medical Diagnostics: How Medical Multimodal Large Language Models Think in Space}

\author{Quoc-Huy Trinh, Xi Ding, Yang Liu, Zhenyue Qin, Xingjian Li, \\
        \bf Gorkem Durak, Halil Ertugrul Aktas, Andrea M. Bejar, Ulas Bagci, Min Xu}

\begin{document}

\maketitle

\begin{abstract}
Visual spatial intelligence is critical for medical image interpretation, yet remains largely unexplored in Multimodal Large Language Models (MLLMs) for 3D imaging. This gap persists due to a systemic lack of datasets featuring structured 3D spatial annotations beyond basic labels. In this study, we introduce an agentic pipeline that autonomously synthesizes spatial visual question-answering (VQA) data by orchestrating computational tools such as volume estimation and bounding boxes extraction with multi-agent collaboration and expert radiologist validation. We present SpatialMed, the first comprehensive benchmark for evaluating 3D spatial intelligence in medical MLLMs, comprising 31,253 question-answer pairs across multiple organs and tumor types. Our evaluations on 24 state-of-the-art MLLMs and extensive analyses reveal that current models lack robust spatial reasoning capabilities for medical imaging.
\url{https://huyquoctrinh.github.io/SpatialMed/}
\end{abstract}

\vspace{-4mm}
\section{Introduction}
Visual-spatial intelligence, the ability to perceive, reason, and manipulate spatial relationships, is a cornerstone of clinical radiology. When interpreting computed tomography (CT) scans, radiologists routinely assess tumor dimensions to stage malignancies, estimate organ volumes to evaluate the organ status, and quantify the relative position of the tumor to plan surgical approaches. These quantitative spatial judgments directly inform treatment decisions: Renal mass less than 4 $cm^{3}$ qualifies as a small renal mass, while a 20\% increase in organ volume could indicate disease progression.

Multimodal Large Language Models (MLLMs) have demonstrated remarkable progress in visual understanding, achieving strong performance on tasks ranging from image captioning to visual question answering in natural scenes~\cite{liu2024llavanext,qwen2vl,glm,internvl,zhu2025internvl3}. Recent work has further extended MLLMs to 2D spatial reasoning, enabling models to describe object positions, compare sizes, and reason about spatial relationships in photographs. In the medical domain, however, current research remains largely restricted to diagnostic classification~\cite{medflamingo,chen2024medblip,chen2024huatuogpt,medmoe,medvlm-r1,shi2024med}, anatomical segmentation~\cite{sammed2d,cheng2023sammed2d,bai2024m3d,baharoon2025rexgroundingct}, and report generation~\cite{ctchat,ctrate1,ctrate2,hamamci2024ct2rep,bassi2025radgpt,hamamcibetter,xin2025med3dvlm}, leaving quantitative visual-spatial reasoning largely unexplored. To our knowledge, no existing work systematically evaluates whether MLLMs can perform grounded spatial reasoning over 3D medical volumes, such as estimating volumetric magnitude or reasoning about inter-structure spatial relationships, both of which are foundational for surgical planning and AI-assisted diagnostics.

\begin{table*}[ht]
\centering
\small
\setlength{\tabcolsep}{4pt}
\renewcommand{\arraystretch}{1.15}
\caption{Comparison of existing 3D CT datasets with \textbf{SpatialMed}. 
Unlike prior work, SpatialMed is the \textit{only} benchmark 
that jointly covers all of spatial reasoning capabilities.}
\label{tab:dataset_comparison}
\resizebox{\textwidth}{!}{
\begin{tabular}{l r r ccc l}
\toprule
\textbf{Dataset} 
& \textbf{CT Scans} 
& \textbf{QA Pairs}
& \textbf{Relative Position}
& \textbf{Quantitative}
& \textbf{Volumetric Relation} 
& \textbf{Primary Tasks} \\
\midrule
\rowcolor{capgray}
\multicolumn{7}{l}{\textit{\small No spatial reasoning support}} \\
INSPECT~\cite{huang2023inspect} 
& 23,248 & 225M 
& \xmark & \xmark & \xmark 
& Diagnosis \\
M3D VQA~\cite{bai2024m3d} 
& 96,170 & 509K 
& \xmark & \xmark & \xmark 
& Diagnosis \\
CT-CHAT~\cite{ctchat} 
& 25,692 & 2.7M 
& \xmark & \xmark & \xmark 
& Diagnosis \\
\addlinespace[2pt]
\rowcolor{capgray}
\multicolumn{7}{l}{\textit{\small Partial spatial reasoning support}} \\
CT-RATE~\cite{hamamci2024developing} 
& 25,692 & 50,186 
& \xmark & \xmark & \xmark 
& Abnormality Detection, Report Generation \\
BIMCV-R~\cite{chen2024bimcv} 
& 2M & 8086 
& \xmark & \xmark & \xmark 
& Text--Image Retrieval \\
3D-RAD~\cite{gai20253d} 
& 16,188 & 136K 
& \cmark & \xmark & \xmark 
& Diagnosis \\
RadImageNet-VQA (CT)~\cite{butsanets2025radimagenet} 
& 750K & 6.75M 
& \xmark & \xmark & \xmark 
& Diagnosis \\
\midrule[1.2pt]
\rowcolor{oursgreen}
\textbf{SpatialMed (Ours)} 
& \textbf{2,375} 
& \textbf{31,253}
& \textcolor{ourstextgreen}{\ding{51}} 
& \textcolor{ourstextgreen}{\ding{51}} 
& \textcolor{ourstextgreen}{\ding{51}} 
& Spatial Question Answering, Volume Estimation \\
\bottomrule
\end{tabular}}
\vspace{-3mm}
\end{table*}

\textbf{The Data Bottleneck.} The primary obstacle to progress is the scarcity of appropriate training and evaluation data. As the demonstration in Table~\ref{tab:dataset_comparison}, existing volumetric radiology datasets provide diagnostic question answering, narrative reports, but systematically lack the structured spatial annotations necessary for quantitative reasoning: \textbf{(1) Precise 3D spatial relationships}, including relative positions, orientations, and topological adjacencies between anatomical structures and lesions; \textbf{(2) Quantitative volumetric measurements}, including volumes, axis dimensions, and cross-sectional areas. Although such information is essential for clinical interpretation, acquiring high-quality spatial annotations is expensive, requiring radiologists to perform case-by-case 3D localization and measurement with consistent anatomical definitions across heterogeneous patient populations.

Alternative approaches face several limitations. Mining spatial information from radiology reports proves unreliable: reports are predominantly narrative, and provide lack of the quantitative relationships and coordinate references needed for spatial understanding. They describe findings qualitatively (for example, ``mass abutting the portal vein'') rather than providing the geometric precision (for example, ``the volume of the brain tumor: 2.3 $cm^{3}$'') required for systematic evaluation.

\textbf{Our approach.} To address this gap, we introduce an agentic pipeline that automatically generates clinically meaningful spatial VQA data from CT scans, the most widely used 3D diagnostic modality. The pipeline employs deterministic spatial computational tools (volume calculators, and bounding boxes extractors) to derive reproducible quantitative measurements from segmentation masks. These measurements are combined with retrieval-augmented generation to produce diverse VQA pairs, which are further validated through multi-agent filtering and review by three board-certified radiologists to ensure clinical validity and non-triviality.

Building upon this pipeline, we present \textbf{SpatialMed}, a comprehensive benchmark comprising approximately 31,253 question-answer pairs across 2,375 CT scans. The benchmark spans 117 anatomical structures and 7 tumor types across multiple organs, including the kidney, pancreas, liver, lung, prostate, and brain. Questions are organized into multiple-choice questions covering directional, extent, volume magnitude, and comparative reasoning, as well as direct volume estimation tasks. Our evaluation of 24 state-of-the-art MLLMs on SpatialMed, encompassing both 2D and 3D model architectures, reveals substantial limitations in their spatial understanding of CT data, with many models performing only marginally above random chance and exhibiting pervasive hallucinations in their reasoning chains. These findings underscore the critical need for targeted advances in spatial reasoning capabilities for medical imaging.

In summary, our contribution includes three folds: 

\begin{itemize}
    \item We design an agentic pipeline that leverages deterministic spatial computational tools, multi-agent collaboration, and radiologist validation to automatically generate high-quality spatial VQA data for CT scans.
    \item  We introduce SpatialMed, the first CT-based benchmark for evaluating spatial intelligence of MLLMs in medical imaging, comprising 31,253 question-answer pairs across 2,375 CT scans covering 117 anatomical structures and 7 tumor types.
    \item  We evaluate 24 MLLMs on SpatialMed, including both 2D multi-view and 3D volumetric models, revealing that current models lack robust spatial reasoning capabilities for medical imaging, with pronounced weaknesses in numerical estimation and reasoning faithfulness.
\end{itemize}

\vspace{-2mm}
\section{Related Works}

\vspace{-2mm}
\textbf{Spatial Reasoning in Vision Language Model.} Spatial reasoning is a fundamental capability of vision–language models for understanding and reasoning about relationships within physical space. To study this problem, several visual question answering (VQA) benchmarks~\cite{yi2019clevrer, gqa, gsr, vissi} have been proposed, which demonstrate strong performance in evaluating spatial reasoning over natural images. However, these benchmarks are limited to the 2D image domain and do not support volumetric data or the medical domain. Although, PRS-Med~\cite{trinh2025prs} tackled relative positional understanding within 2D medical images, the spatial reasoning extends beyond position: attributes like anatomical size and volume are equally critical in clinical and surgical contexts, as it is one of the criteria leads to the treatment and diagnosis of the doctor. 



\begin{figure}[t]
    \centering
    \includegraphics[width=0.9\textwidth]{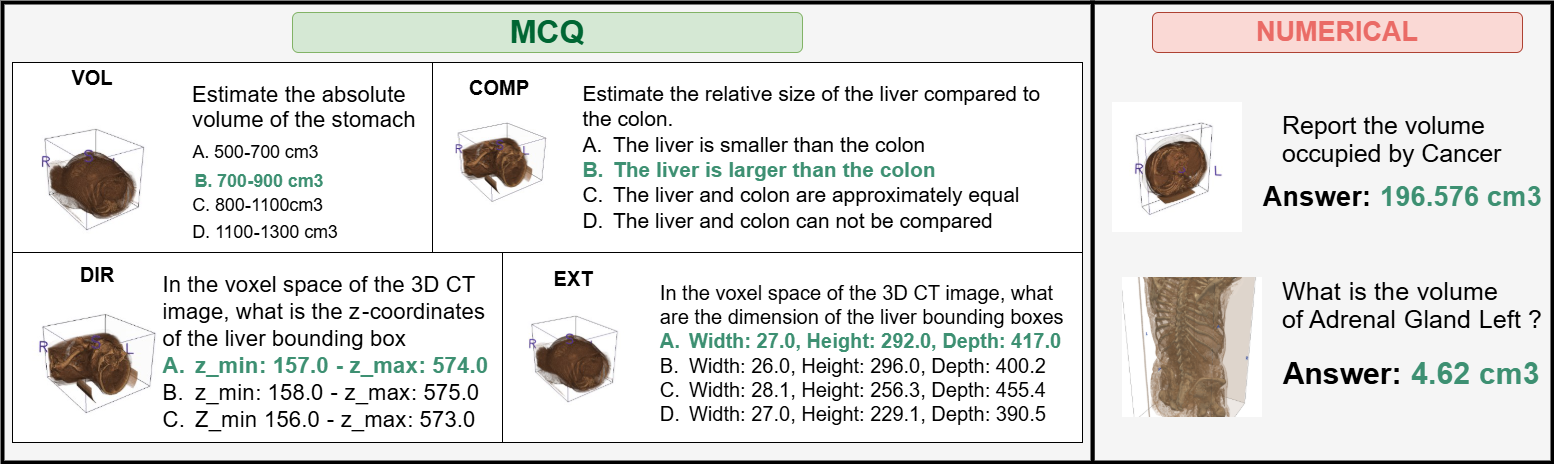}
    \caption{Task demonstrations in the SpatialMed, covering five spatial reasoning questions in two main tasks, including multiple choice questions and quantitative estimation, with corresponding 3D CT visualizations.}
    \label{fig:task_demostration}
    \vspace{-3mm}
\end{figure}

\textbf{3D Medical Imaging Datasets.} Existing 3D medical imaging datasets and benchmarks~\cite{huang2023inspect,bai2024m3d,ctchat,ctrate2,chen2024bimcv,gai20253d,butsanets2025radimagenet} have driven progress in volumetric recognition and report generation. Among these, 3D-RAD~\cite{gai20253d} takes a first step toward spatial evaluation by introducing relative-position questions; however, as shown in Table~\ref{tab:dataset_comparison}, it does not support fine-grained evaluation of relative position, quantitative volume, or inter-organ 3D relations. As a result, in spatial-understanding scenarios, models tend to fall back on diagnostic shortcuts or near-random guessing driven by textual priors, rather than exercising genuine spatial reasoning over the volume.

\textbf{Multimodal Large Language Models in Medical Imaging.} MLLMs have recently achieved strong performance on natural images by jointly reasoning over visual and textual inputs~\cite{liu2024llavanext,glm,qwen3,internvl,zhu2025internvl3,internvl35}, with models such as Qwen3-VL and InternVL3 incorporating explicit spatial reasoning capabilities. In the medical domain, however, most MLLMs operate on 2D slices or weak volumetric proxies~\cite{medflamingo,llavamed,chen2024huatuogpt,medmoe,medvlm-r1,sellergren2025medgemma,unimedvl,lin2025healthgptmedicallargevisionlanguage}, limiting their capacity for three-dimensional spatial reasoning. A growing line of work extends multimodal modeling to 3D medical imaging, including Med-2E3~\cite{shi2024med}, M3D~\cite{bai2024m3d}, BTB3D~\cite{hamamcibetter}, and Med3DVLM~\cite{xin2025med3dvlm}; by integrating dedicated volumetric encoders to capture 3D context. However, these efforts primarily target diagnosis and report generation, evaluating spatial competence only indirectly through downstream task accuracy. Combined with the dataset limitations discussed above, this reveals a fundamental gap between the aggregate task performance of medical MLLMs and their actual ability to reason over 3D spatial structure.

To address aforementioned gaps, we introduce \textbf{SpatialMed}, a benchmark for grounded and faithful spatial reasoning in 3D CT. SpatialMed evaluates core spatial relative position, size, and volume through question-answer pairs explicitly grounded in volumetric evidence. By disentangling answer correctness from reasoning faithfulness, SpatialMed provides a principled framework for benchmarking spatial understanding in medical MLLMs.

\label{sec:related_work}

\vspace{-3mm}
\section{Methods: \textit{SpatialMed}}
\label{sec:method}
\begin{figure}[ht]
    \centering
    \includegraphics[width=1\linewidth, trim=7cm 0cm 0cm 0cm,  
    clip]{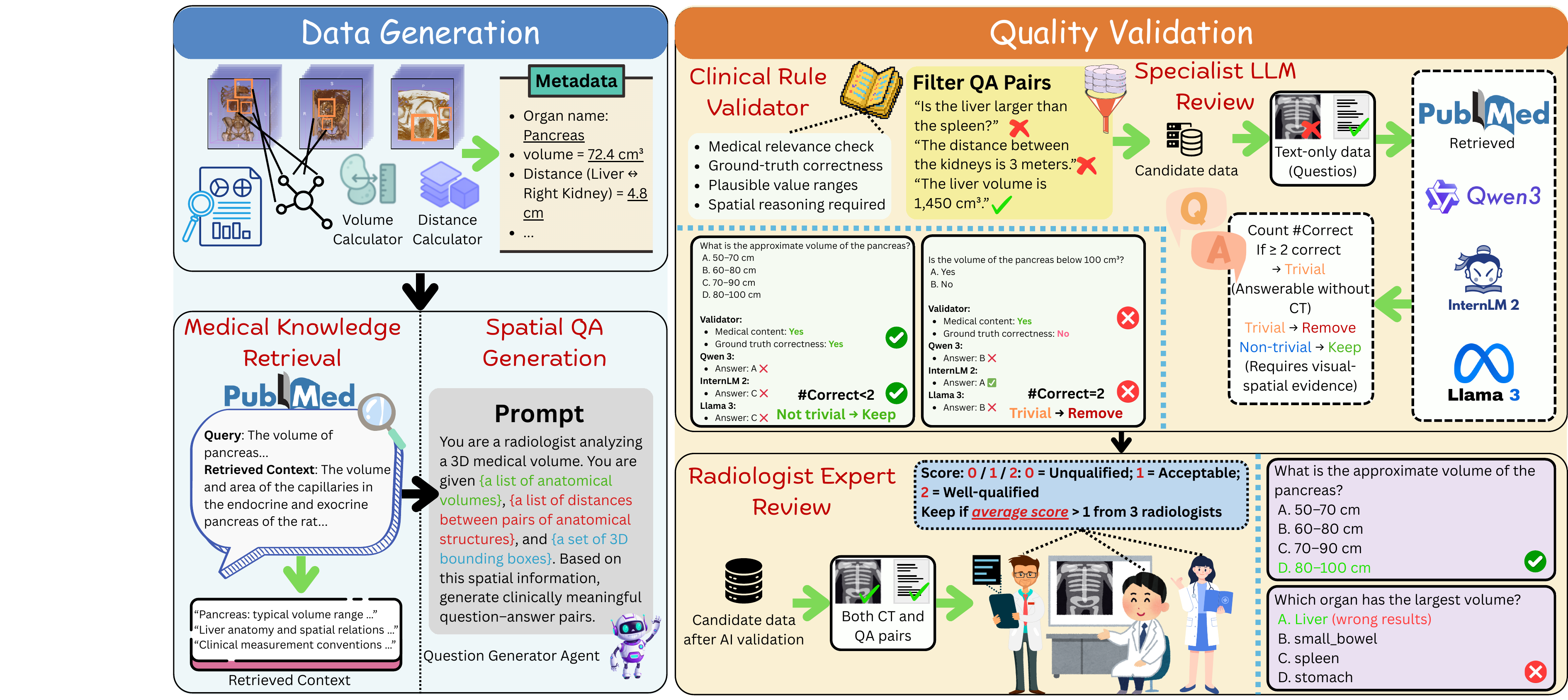}
    \caption{
        \textbf{Overview of three stages from the SpatialMed dataset pipeline.} 
        \textbf{(1) Question--Answer Pair Generation}, where agents produce clinically grounded QA pairs using medical knowledge and spatial analysis tools; \textbf{(2) Data Quality Validation}, in which multiple specialist agents verify medical correctness and the necessity of visual--spatial evidence; and \textbf{(3) Radiologists validation}, where multiple radiologists review participating in review and validate the quality of the dataset.
    }
    \label{fig:data_pipeline}
    \vspace{-2mm}
\end{figure}
\subsection{Data Source}
\label{subsec:datasource}
We construct SpatialMed by curating data from multiple large-scale medical imaging datasets: TotalSegmentator~\cite{wasserthal2023totalsegmentator}, AMOS~\cite{ji2022amos}, the Medical Segmentation Decathlon~\cite{antonelli2022medical}, KiTS~\cite{kits1,kits2}, and BraTS~\cite{brats1,brats2,brats3}. Our dataset comprises 2,375 3D CT scans with corresponding segmentation masks, covering 117 anatomical structures following the TotalSegmentator taxonomy, as well as seven tumor types (including different growing state): renal tumors, pancreatic cancer, liver tumors, prostate tumors, whole tumors, enhancing tumors, and tumor vessel.

Building upon this foundation, \textbf{SpatialMed} encompasses five tasks organized into two categories. The first category consists of multiple-choice questions (MCQ) covering directional reasoning (DIR) (understanding relative anatomical positions), extent/size/
shape reasoning (EXT) (bounding boxes, axis extent, width/height/depth), volume magnitude reasoning (VOL) (numeric values and intervals, including absolute estimation, threshold/range estimation, and numeric calculation), and comparative reasoning (COMP) (comparing entities by magnitude, including pairwise comparison and ranking (largest/smallest)).
The second category requires models to directly estimate volumetric measurements, including absolute volumes, volume ratios, and cross-case comparisons. Examples for each task are demonstrated in Figure~\ref{fig:task_demostration}.

\vspace{-2mm}
\subsection{Agentic Benchmark Construction Pipeline}
\vspace{-2mm}
We introduce a multi-agent pipeline to curate SpatialMed, where specialized tools extract spatial information from medical images, and RAG~\cite{lewis2020retrieval} synthesizes it into QA pairs. One clinical and three specialist agents, with three board-certified radiologists independently confirming clinical validity.

\noindent \textbf{Spatial Computational Tools.} To extract numerical spatial information from 3D CT scans, we introduce two tools: Volume Calculator and 3D Bounding Box Extractor.

\noindent \textit{Volume Calculator.}
To compute the volume of target objects, we employ the 3D segmentation masks. The volume is calculated, as Equation~\ref{equa:vol}, by counting foreground voxels (corresponding to anatomical structures and tumors) and converting this count into a measurement using voxel spacing values in the NIfTI header. Given a segmentation mask 
$M \in \mathbb{R}^{H \times W \times D}$, where foreground voxels are defined as those with values greater than zero, and voxel spacings $(d_x, d_y, d_z)$ in millimeters.
\begin{equation}
V = \sum_{i,j,k} 1\!\left[M_{i,j,k} > 0\right] \cdot d_x d_y d_z \,/\, 1000,
\label{equa:vol}
\end{equation}
where $i,j,k$ denote indices along the height, width, and depth dimensions of the 3D volume, and division by 1000 converts the volume from $\mathrm{mm}^3$ to $\mathrm{cm}^3$. Segmentation masks containing only background voxels are excluded. This voxel-based volume calculator follows standard practice in volumetric medical image segmentation and enables consistent quantitative comparison across datasets with varying spatial resolutions.

\noindent \textit{3D Bounding Box Extractor.}
To extract axis-aligned 3D bounding boxes from segmentation masks, we first identify the set of foreground voxels $\mathcal{S}=\{(i,j,k)\mid M_{i,j,k}>0\}$. The bounding box coordinates are then computed as:
\vspace{-2mm}
\begin{equation}
\begin{aligned}
x_{\min}&=\min_{(i,j,k)\in\mathcal{S}} i,\quad
y_{\min}=\min_{(i,j,k)\in\mathcal{S}} j,\quad
z_{\min}=\min_{(i,j,k)\in\mathcal{S}} k,\\
x_{\max}&=\max_{(i,j,k)\in\mathcal{S}} i,\quad
y_{\max}=\max_{(i,j,k)\in\mathcal{S}} j,\quad
z_{\max}=\max_{(i,j,k)\in\mathcal{S}} k.
\end{aligned}
\end{equation}
In this way, we obtain a tight bounding box enclosing the target anatomy or tumor.
\vspace{-2mm}
\noindent \textbf{Spatial Visual Question-Answer Generation.} All QA pairs are automatically generated by combining voxel-level spatial metadata with five predefined question templates per task, passed to our agent-based framework. To ground generation in medical knowledge, we employ retrieval-augmented generation (RAG)~\cite{lewis2020retrieval} over PubMed~\cite{canese2013pubmed}: given target anatomy or tumor names, the top-5 retrieved documents serve as context for the question annotator agent, enabling medically plausible questions and clinically realistic distractors for multiple-choice items.

Following QA construction in the MCQ, all generated samples are filtered through a validation pipeline consisting of one clinical validation agent and three medical specialist agents. The clinical validation agent applies one-shot prompting to assess each QA pair according to three criteria: \textbf{(1) medical-domain relevance}, requiring anatomies, pathologies, and spatial relations to be clinically meaningful (for example, rejecting ``the volume comparison between the brain tumors and the kidneys'' while accepting ``the volume comparison between the benign brain tumors and maglinant brain tumors''); \textbf{(2) reasonable relative values}, ensuring quantitative attributes such as volume fall within plausible anatomical ranges (for example, rejecting ``the liver volume is 0.01~mm$^3$'' while accepting ``the liver volume is 1{,}450~cm$^3$''); and \textbf{(3) non-triviality}, excluding QA pairs answerable from general medical knowledge alone (for example, ``Is the liver larger than the spleen?''), while retaining those requiring CT image-grounded spatial reasoning.
Subsequently, three medical specialist agents powered by InternLM 2~\cite{cai2024internlm2}, Qwen-3~\cite{qwen3}, and Llama-3~\cite{llama3} attempt to answer each question using only the RAG tool~\cite{lewis2020retrieval}, without access to volumetric images.
If at least two specialists succeed, the sample is deemed trivial and discarded; otherwise it is retained. After this multi-agent validation and filtering process, the dataset is reduced from 30{,}799 automatically generated samples to 10{,}487 high-quality spatial reasoning QA pairs.

For visual input, 3D-capable MLLMs receive the full CT volume. While in 2D, since models can not handle 3D inputs, three orthogonal cross-sectional views (axial, coronal, sagittal along the $z$, $y$, $x$ axes) are ingested, each sliced at the centroid of the target anatomy's segmentation mask. Crucially, each slice preserves the full body cross-section rather than being cropped around the target, so the model must still localize the structure and reason about its spatial context. This mask-guided selection is necessary rather than an unfair advantage: without it, the target may be absent from the slices entirely, making spatial reasoning impossible. Since 3D and 2D models receive inherently different representations, the protocol is designed to explore the spatial understanding capability of the 2D MLLMs, and assess the 3D spatial understanding of 3D models.

\begin{figure}[t]
    \centering
    \includegraphics[width=0.23\columnwidth]{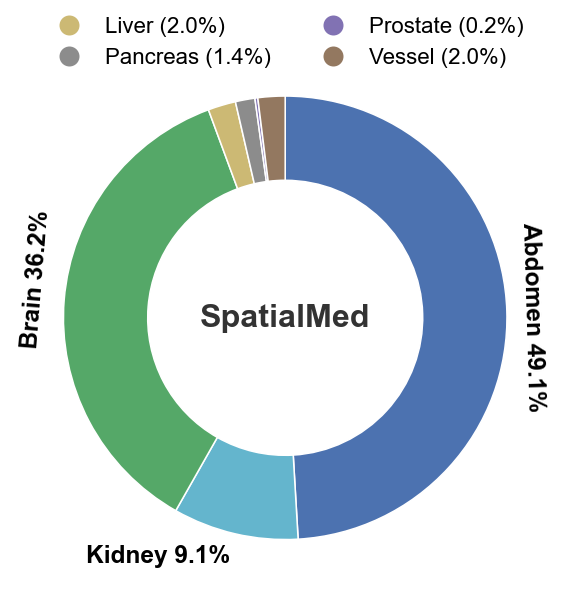}
    \hfill
    \includegraphics[width=0.44\columnwidth]{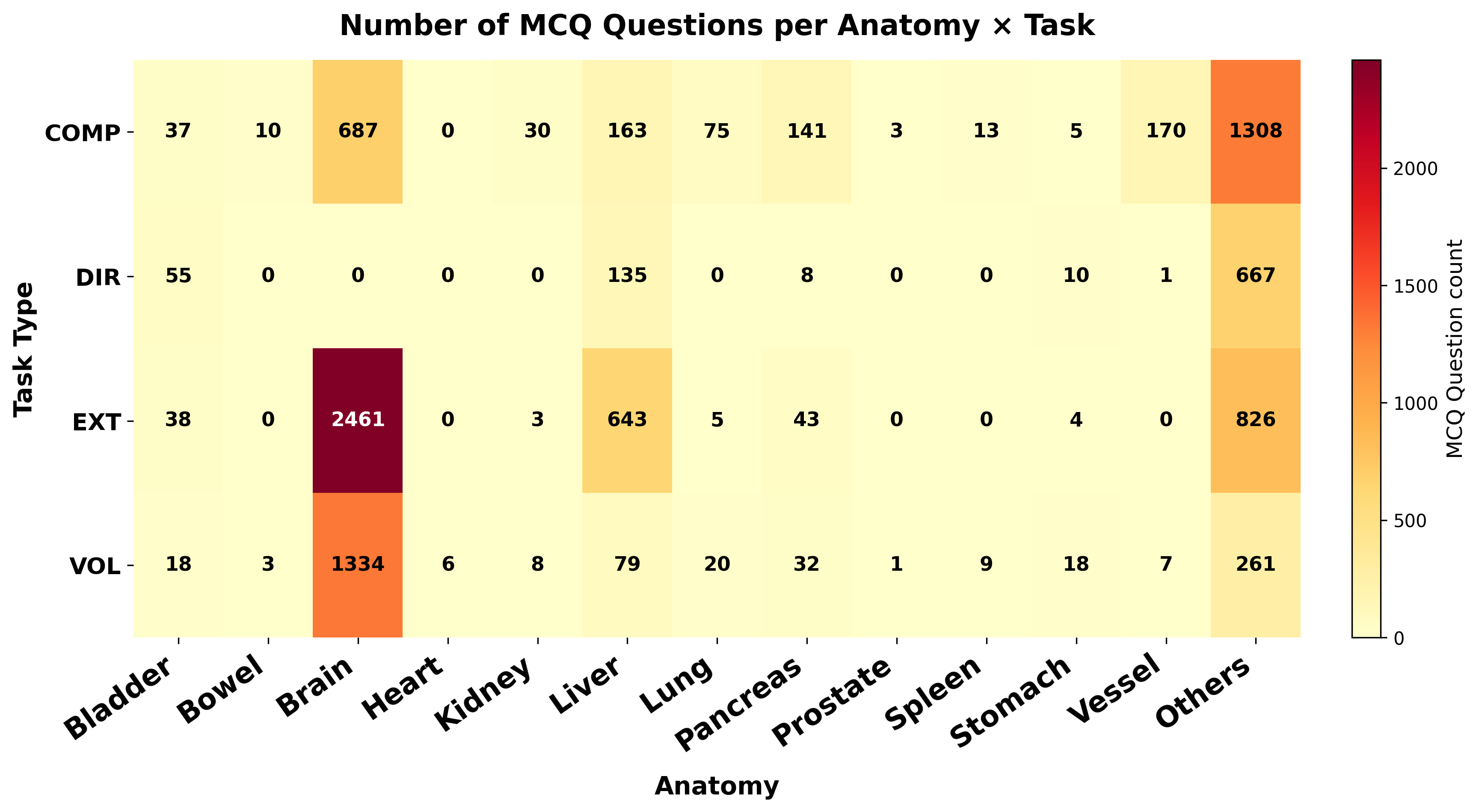}
    \hfill
    \includegraphics[width=0.23\columnwidth]{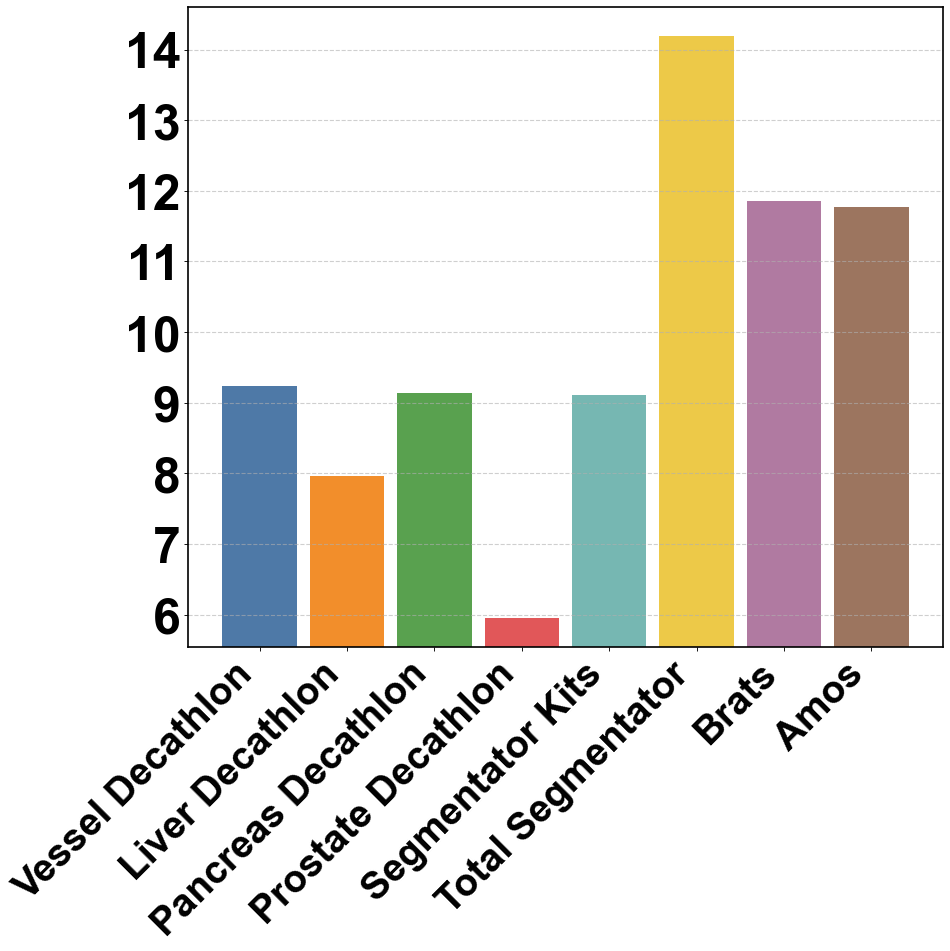}
    \caption{\textbf{Benchmark Statistics.} \textbf{Left:} Distribution of annotated anatomical regions in the MCQ task. \textbf{Middle:} Coverage of top anatomies across the five task types. \textbf{Right:} Dataset distribution across the volume task, where the y-axis is shown on a \textbf{log$_2$ scale}.}
    \label{fig:benchmark_stats}
    \vspace{-5mm}
\end{figure}

\textbf{Expert-in-the-loop Quality Review.} 
For the MCQ, to ensure the benchmark's clinical integrity and mitigate "hallucination artifacts" common in LLM-generated content, we implemented a rigorous, multi-stage validation phase involving three board-certified radiologists. This process refined the initial pool of \textbf{10,487} VQA pairs down to a final, high-fidelity set of \textbf{8864} samples. For the volume calculation tasks, the initial 27,906 samples are filtered down carefully to 22,389 by three board-certified radiologists, and all of the incorrect volume samples are removed.

Each radiologist independently audited the full 3D volumetric scans alongside the generated QA pairs, assigning a quality score based on three stringent dimensions:
\textbf{(1) Clinical Logicality} -- Evaluators rejected any queries that, while spatially correct, were medically nonsensical or lacked diagnostic utility, ensuring every question mirrors a real-world clinical scenario; \textbf{(2) Quantitative Consistency} -- Answers were verified against the imaging evidence to ensure that volumetric statistic remained within plausible anatomical ranges; \textbf{(3) Exhaustive Mutuality}-- For comparative tasks, radiologists ensured that choice sets (for example, volume change or size ranking) were both mutually exclusive and exhaustive, preventing "lucky guesses" through flawed distractor logic. Regarding the volume calculation tasks, to keep the workload tractable, each of the radiologists reviews one-third of the candidate pool and removes samples whose source segmentation is unreliable.

Scores are determined as:  0 for unqualified, 1 for acceptable, and 2 for well-qualified QA pairs. Final inclusion in \textbf{SpatialMed} required an average score from three radiologists greater than 1.0, effectively filtering out samples where the spatial evidence is ambiguous or the reasoning path is non-interpretable by a human expert. This expert-driven bottleneck transforms \textbf{SpatialMed} from a synthetic dataset into a validated clinical instrument for measuring and attributing MLLM reliability, which makes the dataset more realistic and be aligned with the scope from the medical application.
\vspace{-4mm}
\subsection{Dataset Distribution}
The overall dataset distribution is illustrated in
Figure~\ref{fig:benchmark_stats}. The MCQ task is dominated by the abdomen (3997) and the brain (3209), while the volume task is mainly from TotalSegmentator (18,675) and BraTS (3,714). The number of organs in the abdominal regions contributes to the diversity in the anatomical and organs of the dataset.

\begin{table*}[t]
\centering
\caption{\textbf{Spatial Reasoning Benchmark.} Comparison of 2D vision--language models on spatial reasoning tasks. The best result for each category is highlighted in \textbf{bold}. For 2D models, the upper block corresponds to \emph{non-medical pretraining}, while the lower block uses \emph{medical-domain pretraining}. \textbf{DIR} denotes \emph{Directional Reasoning}, \textbf{EXT} denotes \emph{Extent / Size / Shape Reasoning}, \textbf{VOL} denotes \emph{Volume Magnitude Reasoning}, and \textbf{COMP} denotes \emph{Comparative Reasoning.}}
\label{tab:med_vlm_grouped}
\setlength{\tabcolsep}{3.0pt}
\renewcommand{\arraystretch}{1.1}
\small
\resizebox{\textwidth}{!}{
\begin{tabular}{l l l l c| c c c c c}
\toprule
\multirow{2}{*}{\textbf{Method}} &
\multirow{2}{*}{\textbf{Size}} &
\multirow{2}{*}{\textbf{Vision backbone}} &
\multirow{2}{*}{\textbf{Language backbone}} &
\multicolumn{5}{c}{\textbf{MCQ (\%)}} &
\multirow{2}{*}{\textbf{Volume}} \\
\cmidrule(lr){5-9}
 &  &  & 
 & \textbf{AVG} & \textbf{DIR} & \textbf{EXT} & \textbf{VOL} & \textbf{COMP} &  \\
\midrule
Random
 &  &  & 
 & 25.00 & 25.00 & 25.00 & 25.00 & 25.00 &  \\
\midrule
\multicolumn{10}{l}{\textbf{Private Models}} \\
\cmidrule(lr){1-10}
GPT-5.4 (2026)~\cite{singh2025openai}
 &  &  & 
 & 39.23 & 42.48 & 39.36 & \textbf{36.26} & 40.03 & \textbf{41.06} \\
\rowcolor{green!15}
Gemini-2.5-flash (2025)~\cite{comanici2025gemini}
 &  &  & 
 & \textbf{43.11} & \textbf{51.82} & \textbf{47.05} & 31.93 & \textbf{42.01} & 37.84 \\
\midrule
\multicolumn{10}{l}{\textbf{2D Models (Non-medical Pretraining)}} \\
\cmidrule(lr){1-10}
LLaVA-Next (2024)~\cite{liu2024llavanext}
 & 7B & SigLIP-400M & Mistral-7B
 & 48.00 & \textbf{56.83} & 51.43 & 39.03 & 46.19 & 21.35 \\
GLM-4.1V-9B (2025)~\cite{glm}
 & 9B & AIMv2-Huge & GLM-4.1
 & 28.08 & 22.37 & 25.49 & 31.11 & 31.77 & 22.93 \\
GLM-4.6V-Flash (2025)~\cite{zeng2025glm}
 & 9B & AIMv2-Huge & GLM-4.6
 & 35.85 & 20.68 & 42.50 & 27.98 & 36.34 & 10.90 \\
Qwen3-VL 4B (2025)~\cite{bai2025qwen3}
 & 4B & SigLIP2-Large & Qwen3-4B
 & 50.59 & 55.26 & 45.80 & \textbf{53.11} & 54.53 & 34.14 \\
Qwen3-VL 8B (2025)~\cite{bai2025qwen3}
 & 8B & SigLIP2-SO-400M & Qwen3-8B
 & 43.28 & 34.82 & 39.28 & 48.72 & 48.32 & 20.72 \\
InternVL3 8B (2025)~\cite{zhu2025internvl3}
 & 8B & InternViT-300M & Qwen2.5-7B
 & 35.44 & 43.29 & 25.41 & 36.47 & 47.32 & 28.69 \\
InternVL3 9B (2025)~\cite{zhu2025internvl3}
 & 9B & InternViT-300M & InternLM3-8B
 & 44.38 & 52.00 & 39.38 & 37.15 & 54.53 & 35.92 \\
Qwen3.5 9B (2026)~\cite{qwen35}
 & 9B & Native ViT & Qwen3.5-9B
 & 41.08 & 50.54 & 51.00 & 26.72 & 32.97 & 30.89 \\
InternVL3.5 14B (2026)~\cite{internvl35}
 & 15B & InternViT-300M & Qwen3-14B
 & 46.03 & 46.55 & 45.77 & 31.17 & 56.69 & 11.00 \\
\rowcolor{green!15}
InternVL3.5 30B-A3B (2026)~\cite{internvl35}
 & 31B (A3B) & InternViT-300M & GPT-OSS
 & \textbf{52.73} & 51.63 & \textbf{52.85} & 40.28 & \textbf{61.66} & \textbf{36.99} \\
\midrule
\multicolumn{10}{l}{\textbf{2D Models (Medical Pretraining)}} \\
\cmidrule(lr){1-10}
MedFlamingo (2023)~\cite{medflamingo}
 & 9B & CLIP ViT-L/14 & LLaMA-7B
 & 38.03 & 36.15 & 35.82 & 38.69 & 41.55 & 13.93 \\
LLaVA-Med (2024)~\cite{llavamed}
 & 7B & CLIP ViT-L/14 & Mistral-7B
 & 56.68 & \textbf{73.76} & 54.12 & 55.27 & 55.89 & 15.17 \\
HuatuoGPT-Vision (2024)~\cite{chen2024huatuogpt}
 & 7B & CLIP ViT-L/14 & Qwen2-7B
 & 1.66 & 1.33 & 1.35 & 2.51 & 1.64 & NaN \\
MedMoE (2024)~\cite{medmoe}
 & 4B & CLIP ViT-L/14 & Phi-2-2.7B
 & 21.23 & 22.13 & 21.39 & 27.64 & 16.19 & 16.45 \\
Med-VLM R1 (2025)~\cite{medvlm-r1}
 & 8B & Qwen2-VL's ViT & Qwen2-VL-2B
 & 6.14 & 4.84 & 3.65 & 7.12 & 9.66 & NaN \\
Med-Gemma 4B (2025)~\cite{sellergren2025medgemma}
 & 4B & SigLIP-400M & Gemma 3 4B
 & 52.30 & 54.66 & 51.61 & 46.55 & 56.61 & 25.69 \\
UniMedVL (2025)~\cite{unimedvl}
 & 14B & ViT-B/16 & BioMedBERT
 & 41.94 & 37.12 & 35.39 & 39.54 & 55.17 & 9.47 \\
\rowcolor{green!15}
HealthGPT-XL32 (2025)~\cite{lin2025healthgptmedicallargevisionlanguage}
 & 32B & CLIP ViT-L/14 & Qwen 2.5 32B
 & \textbf{62.44} & 57.80 & \textbf{69.97} & \textbf{55.38} & \textbf{57.53} & \textbf{31.11} \\
\bottomrule
\end{tabular}
}
\vspace{-4mm}
\end{table*}

\vspace{-2mm}
\begin{table*}[!ht]
\centering
\caption{\textbf{Spatial Reasoning Benchmark on 3D Models.} Comparison of 3D vision--language models on spatial reasoning tasks. The best result for each category is highlighted in \textbf{bold}. \textbf{DIR} denotes \emph{Directional Reasoning}, \textbf{EXT} denotes \emph{Extent / Size / Shape Reasoning}, \textbf{VOL} denotes \emph{Volume Magnitude Reasoning}, and \textbf{COMP} denotes \emph{Comparative Reasoning}.}
\label{tab:med_vlm_3d}
\setlength{\tabcolsep}{3.0pt}
\renewcommand{\arraystretch}{1.1}
\small
\resizebox{\textwidth}{!}{
\begin{tabular}{l l l l c| c c c c c}
\toprule
\multirow{2}{*}{\textbf{Method}} &
\multirow{2}{*}{\textbf{Size}} &
\multirow{2}{*}{\textbf{Vision backbone}} &
\multirow{2}{*}{\textbf{Language backbone}} &
\multicolumn{5}{c}{\textbf{MCQ (\%)}} &
\multirow{2}{*}{\textbf{Volume}} \\
\cmidrule(lr){5-9}
 &  &  & 
 & \textbf{AVG} & \textbf{DIR} & \textbf{EXT} & \textbf{VOL} & \textbf{COMP} &  \\
\midrule

Med-2E3 (2024)~\cite{shi2024med}
 & 3B & M3D-CLIP+SigLIP & Qwen2.5-3B-Instruct
 & 41.19 & 34.10 & 37.48 & \textbf{45.41} & 46.19 & \textbf{31.58} \\
M3D (2024)~\cite{bai2024m3d}
 & 7B & M3D-CLIP & LLaMA-2-7B
 & 21.92 & 0.12 & 27.13 & 33.22 & 13.30 & NaN \\

  BTB3D (2025)~\cite{hamamcibetter}
 & 8B & CT-ViT & LLaMA-3.1-8B-Instruct
 & 31.86 & 39.66 & 35.58 & 22.51 & 30.21 & 18.62 \\
  \rowcolor{green!15}
 Med3DVLM (2025)~\cite{xin2025med3dvlm}
 & 7B & DCFormer-SigLIP & Qwen2.5-7B-Instruct
 & \textbf{53.80} & \textbf{71.10} & \textbf{55.23} & 41.37 & \textbf{54.65} & 21.90 \\

\bottomrule
\end{tabular}
}
\end{table*}

\vspace{-3mm}
\section{Experiments and Results}
\vspace{-1mm}
\subsection{Experimental Setup}
\textbf{Benchmark Model.} We conduct a comprehensive evaluation of 24 MLLMs across domains, covering a wide range of model sizes, architectures, and training recipes. Regarding the private models, we do the benchmark on GPT-5.4~\cite{singh2025openai}, and Gemini-2.5-Flash~\cite{comanici2025gemini}. For open-source general-domain models, we evaluate LLaVA-Next~\cite{liu2024llavanext}, GLM-4.1V~\cite{glm}, GLM-4.6V~\cite{zeng2025glm}, Qwen3-VL~\cite{bai2025qwen3}, and InternVL 3~\cite{zhu2025internvl3}, InternVL 3.5~\cite{internvl35}. For models pretrained or fine-tuned on medical datasets, we include MedFlamingo~\cite{medflamingo}, LLaVA-Med~\cite{llavamed}, HuatuoGPT-Vision~\cite{chen2024huatuogpt}, MedMoE~\cite{medmoe}, Med-VLM R1~\cite{medvlm-r1}, Med-Gemma~\cite{sellergren2025medgemma}, UniMedVL~\cite{unimedvl}, HealthGPT-XL32~\cite{lin2025healthgptmedicallargevisionlanguage}. Beyond 2D image–based MLLMs, we evaluate 3D medical VLM, including Med-2E3~\cite{shi2024med}, M3D~\cite{bai2024m3d}, BTB3D~\cite{hamamcibetter}, and Med3DVLM~\cite{xin2025med3dvlm}. All benchmarks are conducted under a zero-shot evaluation setting. To ensure fairness and reproducibility, we apply greedy decoding consistently across all evaluated models.

\textbf{Metric Design.} For MCQ tasks (Section~\ref{subsec:datasource}), we adopt Accuracy as the primary evaluation metric, computed via exact matching between the predicted option and the ground-truth answer. For numerical estimation tasks, we employ the Mean Relative Accuracy (MRA)~\cite{vissi}, which assesses prediction quality under varying tolerance levels. As defined in Equation~\ref{eqa:mra}, we average the accuracy across strict confident thresholds $C = \{0.01, 0.02, \ldots, 0.1\}$, following Vissi~\cite{vissi}. MRA measures the proportion of predictions whose relative error falls within each confidence bound and averages these proportions across all thresholds. This formulation captures both absolute correctness and numerical proximity, providing a fine-grained evaluation of quantitative reasoning performance.
\vspace{-2mm}
\begin{equation}
    \mathrm{MRA} = 100 \times \frac{1}{|\theta|} \sum_{\theta \in C} 1\!\left( \frac{|\hat{y} - y|}{y} < 1 - \theta \right),
    \label{eqa:mra}
\end{equation}

Where $\hat{y}$ denotes the predicted value, $y$ is the ground truth, and $\theta$ is the confidence threshold. The overall model performance is computed by averaging the results across the five tasks.

\begin{figure*}[t]
    \vspace{-2mm}
    \centering
    
    \begin{subfigure}{0.3\textwidth}
        \centering
        \includegraphics[width=\linewidth]{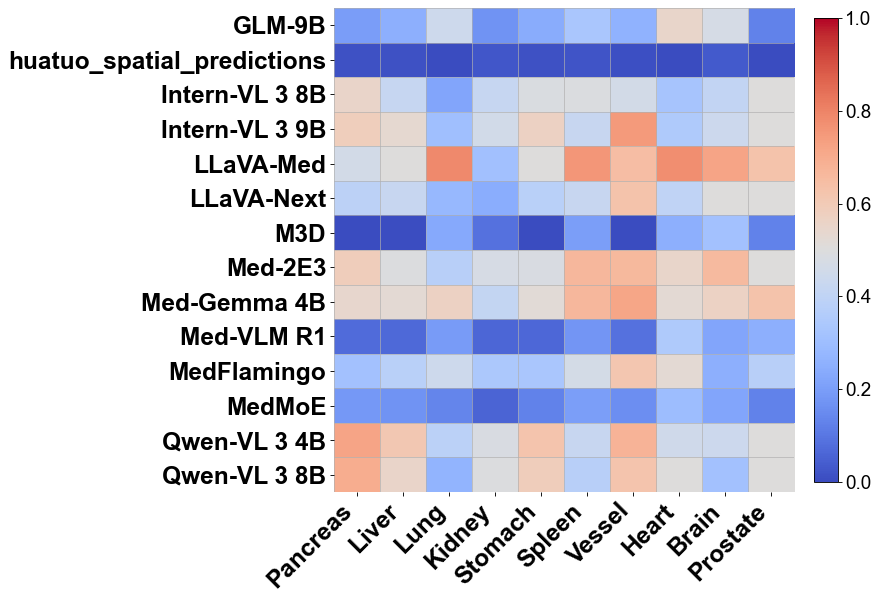}
        \caption{Per-organ accuracy.}
        \label{fig:organ_accuracy_heatmap}
    \end{subfigure}
    \hfill
    \begin{subfigure}{0.3\textwidth}
        \centering
        \includegraphics[width=\linewidth]{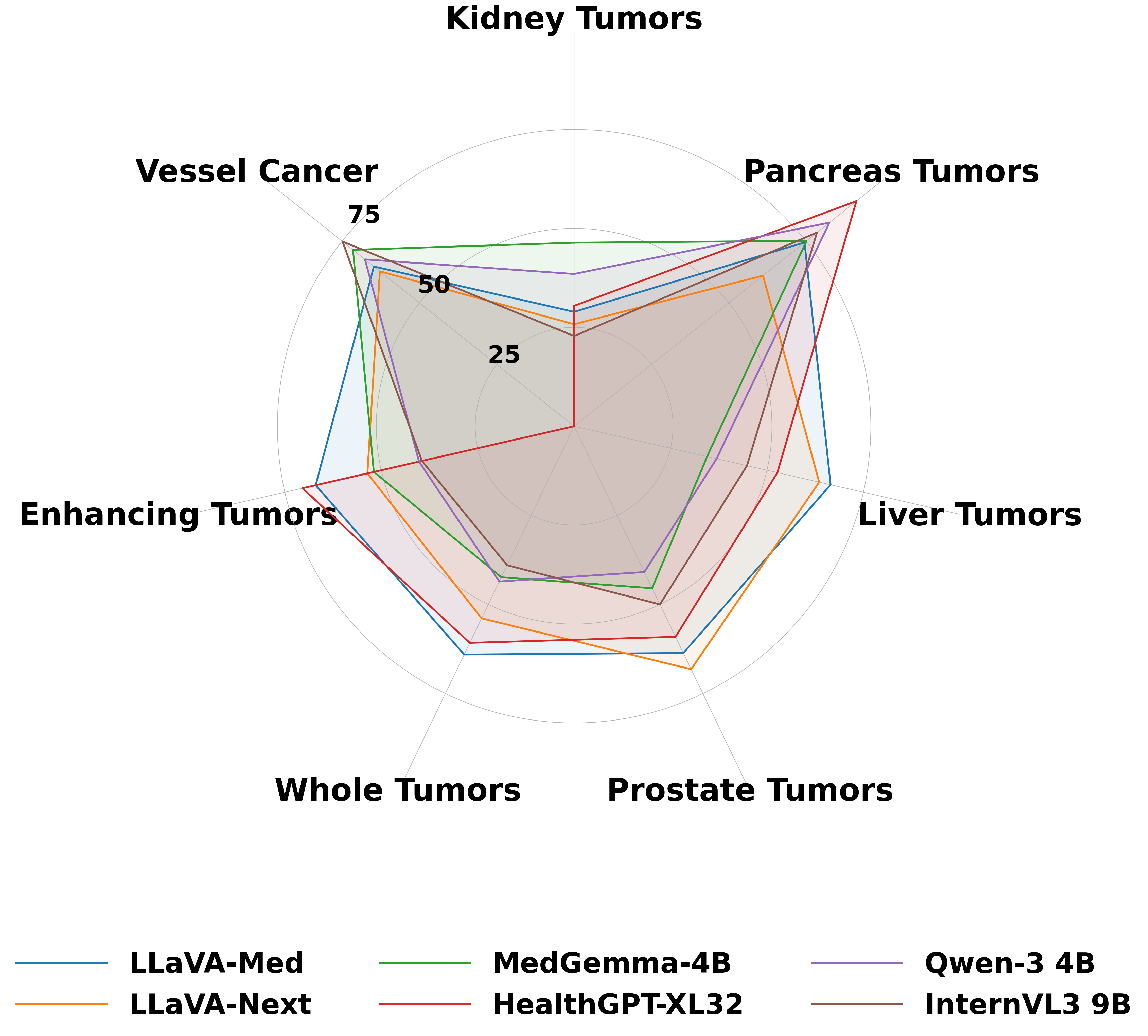}
        \caption{Tumor-wise accuracy.}
        \label{fig:tumor_organ_radar}
    \end{subfigure}
    \hfill
    \begin{subfigure}{0.36\textwidth}
        \centering
        \includegraphics[width=\linewidth]{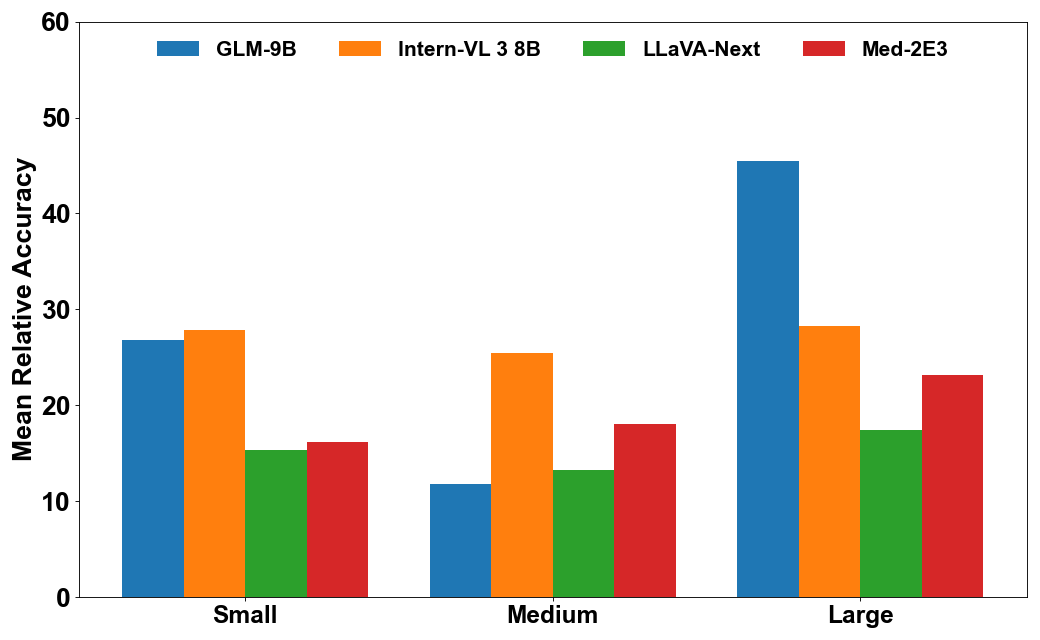}
        \caption{Volume-bucketed performance.}
        \label{fig:volume_bucket}
    \end{subfigure}
    
    \caption{\textbf{Fine-grained performance analysis across anatomical structures, tumor types, and volume scales.}
    (a) Per-organ accuracy across models in the MCQ task.
    (b) Tumor-wise accuracy across selected models.
    (c) Performance stratified by anatomical volume buckets using Mean Relative Accuracy.}
    \label{fig:MCQ_analysis}
    \vspace{-3mm}
\end{figure*}


\vspace{-3mm}
\subsection{Evaluation on SpatialMed}
Table~\ref{tab:med_vlm_grouped} presents evaluation by aforementioned metrics across multiple model categories, including 2D models (with/without medical pretraining), and Table~\ref{tab:med_vlm_3d} illustrates the results of 3D models. Models cannot produce numerical outputs; the corresponding results are reported as NaN.

\textbf{Multiple Choice results.} In the 2D domain, performance on the MCQ tasks spans in a wide range, and three key findings have emerged.\textbf{(1)~Medical pretraining is necessary but insufficient:}
The two best-performing models are both medically pretrained, including HealthGPT-XL32~\cite{lin2025healthgptmedicallargevisionlanguage}
(62.44\%) and LLaVA-Med~\cite{llavamed} (56.68), and Med-Gemma 4B~\cite{sellergren2025medgemma}
(52.30\%) also outperforms most non-medical models at a fraction of the parameter count.
However, medical pretraining alone is not sufficient: HuatuoGPT-Vision~\cite{chen2024huatuogpt}
(1.66\%) and Med-VLM R1~\cite{medvlm-r1} (6.14\%) collapse far below the 25\% random baseline,
indicating that these models have a deficiency in the medical spatial understanding.
Non-medical models such as InternVL3.5 30B-A3B~\cite{internvl35} (52.73\%) and
Qwen3-VL 4B~\cite{bai2025qwen3} (50.59\%) remain competitive through architectural
strength alone, but do not surpass the best medical systems.
\textbf{(2)~General-domain scale still limits with the medical spatial understanding:} GPT-5.4~\cite{singh2025openai} and Gemini 2.5 Flash~\cite{comanici2025gemini} trail the strongest close-source systems on this benchmark, indicating that general-domain capability does not automatically compensate for the lack of domain-specific spatial grounding.
\textbf{(3)~2D models are deficient for 3D understanding:} no single 2D model dominates every subtask, LLaVA-Med~\cite{llavamed} leads directional reasoning with a striking 73.76\% DIR accuracy, HealthGPT-XL32~\cite{lin2025healthgptmedicallargevisionlanguage} leads extent and volume-magnitude estimation, and InternVL3.5 30B-A3B~\cite{internvl35} leads comparative reasoning, suggesting that these sub-capabilities trade off against one another as the backbone or training recipe changes. In addition, a fundamental bottleneck remains architectural: current 2D vision encoders cannot natively encode long sequences of contiguous slices or 3D topological structures needed to convey volumetric context to the language model, which is reflected in universally low Volume MRA scores.


Among 3D models (Table~\ref{tab:med_vlm_3d}), two insights emerge.
\textbf{(1)~Volumetric input enables strong directional reasoning but does not
guarantee volume estimation superiority:} Med3DVLM~\cite{xin2025med3dvlm} leads
most subtasks with 71.10\% DIR, comparable to the best 2D result, yet the best
3D Volume MRA (Med-2E3~\cite{shi2024med}, 31.58\%) barely matches the best 2D
medical result (HealthGPT-XL32, 31.11\%) and still trails GPT-5.4 (41.06\%),
implying that volumetric input alone does not translate into better magnitude
reasoning without task-aligned supervision. This gap highlights the need for
improved alignment methods between textual and 3D representations that can
effectively convey spatial and topological concepts to the language model,
enabling stronger 3D spatial understanding.
\textbf{(2)~Training data composition is the primary bottleneck:} despite
promising results, most 3D models derive their supervision from
M3D~\cite{bai2024m3d} and CT-RATE~\cite{hamamci2024developing}, which
predominantly cover diagnostic question answering and report generation,
limiting their ability to generalize to spatial reasoning tasks.

\textbf{Organ and tumor analysis.}
The per-organ heatmap in Figure~\ref{fig:organ_accuracy_heatmap} reveals substantial heterogeneity, with model ranking unstable across organs. Weaker models (HuatuoGPT-Vision~\cite{chen2024huatuogpt}, M3D~\cite{bai2024m3d}, Med-VLM R1~\cite{medvlm-r1}, MedMoE~\cite{medmoe}, MedFlamingo~\cite{medflamingo}) remain uniformly low, suggesting that they fail to localize the referenced structure. Stronger models (LLaVA-Med~\cite{llavamed}, Med-Gemma 4B~\cite{sellergren2025medgemma}) yield more balanced heatmaps but still exhibit clear organ-conditioned gaps. CT-grounded spatial reasoning is thus bounded not only by reasoning operators such as organ boundaries, but also by upstream perception of local context, and multi-view projection stability—motivating organ-conditioned reporting over a single average.

Figure~\ref{fig:tumor_organ_radar} isolates tumor-related MCQ questions across the kidney, pancreas, liver, prostate, whole and enhancing tumors, vessel cancer, and edema. Leading models (LLaVA-Med~\cite{llavamed}, Med-2E3~\cite{shi2024med}, InternVL3 9B~\cite{zhu2025internvl3}, Qwen3-VL 4B~\cite{qwen3}) trace broadly similar shapes but diverge on specific axes, indicating that gains are non-uniform across tumor categories. Near complex boundaries, small localization errors flip relative-distance or adjacency judgments, so stronger tumor recognition alone does not guarantee robust spatial reasoning. Future training should explicitly couple lesion localization with geometric-relation supervision.
\textbf{Volume estimation results.} Volume estimation remains challenging for most models. GPT-5.4~\cite{singh2025openai} achieves the best MRA at 41.06, followed by Gemini 2.5 Flash~\cite{comanici2025gemini}, InternVL3.5 30B-A3B~\cite{internvl35}, InternVL3 9B~\cite{zhu2025internvl3}, Qwen3-VL 4B~\cite{qwen3}. In the 3D model, the Med-2E3~\cite{shi2024med} shows the best performance compared to other methods. Additionally, the top 3D score barely matches the best 2D medical result (HealthGPT-XL32~\cite{lin2025healthgptmedicallargevisionlanguage}) and sits below several open-source 2D baselines, confirming that 3D volumetric input alone does not yet translate into better magnitude estimation without explicit quantitative supervision. Scale is also insufficient on its own: Qwen3-VL 8B~\cite{qwen3} regresses sharply from its 4B counterpart, and larger variants such as InternVL3.5 14B~\cite{internvl35} and GLM-4.6V-Flash~\cite{zeng2025glm} likewise fall to low double-digit MRA. Because neighbouring variants differ in both vision and language backbones, these regressions reflect visual feature granularity and instruction recipe rather than language scale alone. Medical pretraining also fails to help consistently: LLaVA-Med~\cite{llavamed}, MedFlamingo~\cite{medflamingo}, MedMoE~\cite{medmoe}, and UniMedVL~\cite{unimedvl} all lag several general-domain models, suggesting that medical tuning prioritizes recognition over measurement. Private systems, in contrast, achieve the best volume MRA even though their MCQ averages sit mid-pack, implying that general scale and calibration quality can partially compensate for the absence of medical-specific visual priors in magnitude estimation.

Three models (HuatuoGPT-Vision~\cite{chen2024huatuogpt}, Med-VLM R1~\cite{medvlm-r1}, and M3D~\cite{bai2024m3d}) return NaN for volume estimation, exposing a failure to handle the numerical estimation tasks. This is consistent with known miscalibration in instruction-tuned models~\cite{shen2024thermometer} and motivates explicit numeric supervision, unit-aware training, and structured output constraints for quantitative clinical tasks. Figure~\ref{fig:volume_bucket} further stratifies MRA by anatomical volume range for four representative models. It shows volume-dependent performance that is not uniformly monotonic and remains error-sensitive because minor absolute mistakes cause large relative deviations, for example, GLM-9B~\cite{glm,zeng2025glm} peaks on Large lesions and dips on Medium, while InternVL3 8B~\cite{zhu2025internvl3} is nearly flat across buckets and LLaVA-Next~\cite{liu2024llavanext} trails in every bucket. 

\vspace{-3mm}
\subsection{Failure analysis}
\vspace{-1mm}
Following \cite{ribeiro2016should} and \cite{vissi} setups, we randomly sample 50 failure cases in reasoning with the chain-of-thought setup for each task to assess the limitations in the MLLMs' spatial understanding.


\begin{figure}[!ht]
\vspace{-2mm}
    \centering
    
    \begin{subfigure}{0.45\linewidth}
        \centering
        \includegraphics[width=\linewidth]{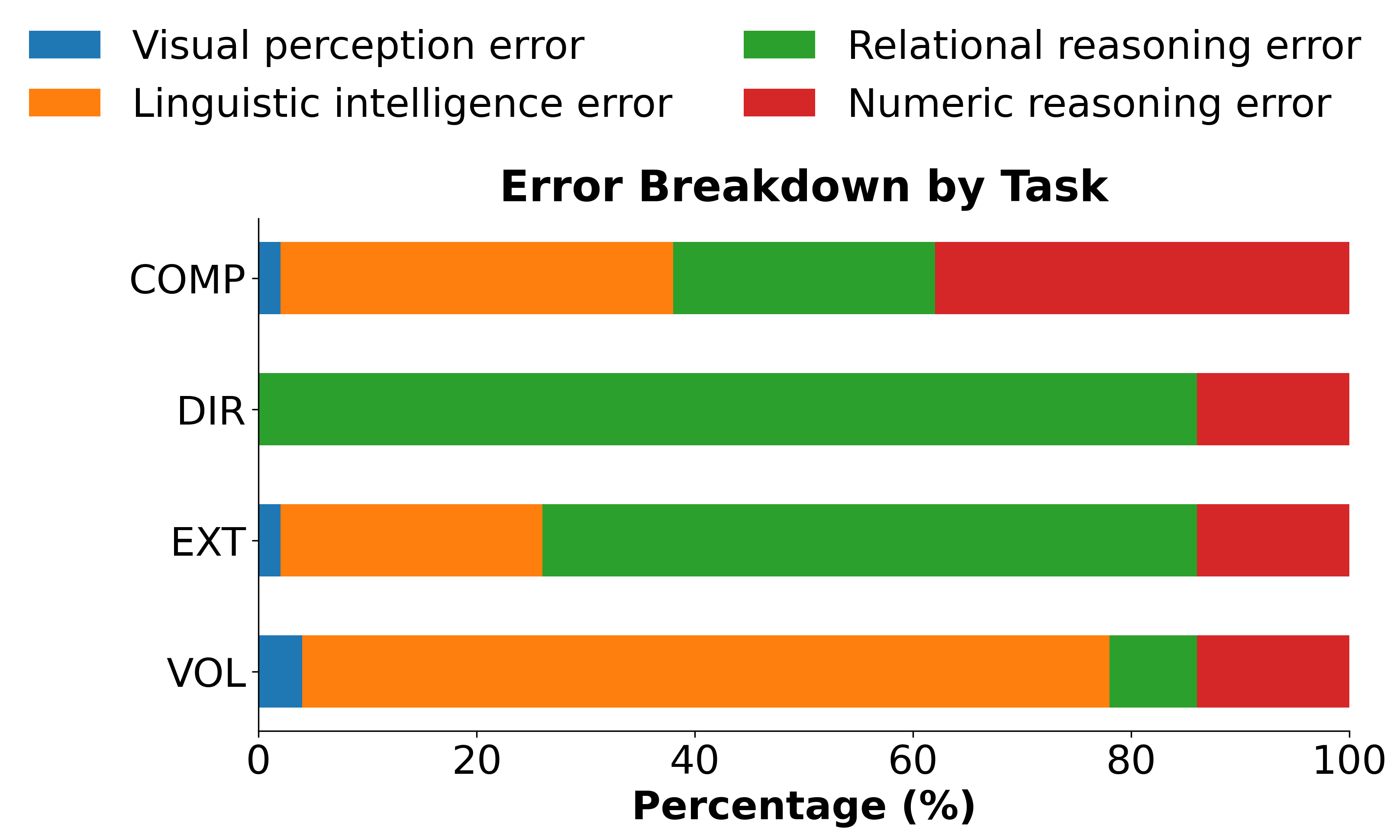}
        \caption{Human analysis of reasoning error types.}
        \label{fig:error_analysis}
    \end{subfigure}
    \hfill
    \begin{subfigure}{0.32\linewidth}
        \centering
        \includegraphics[width=\linewidth]{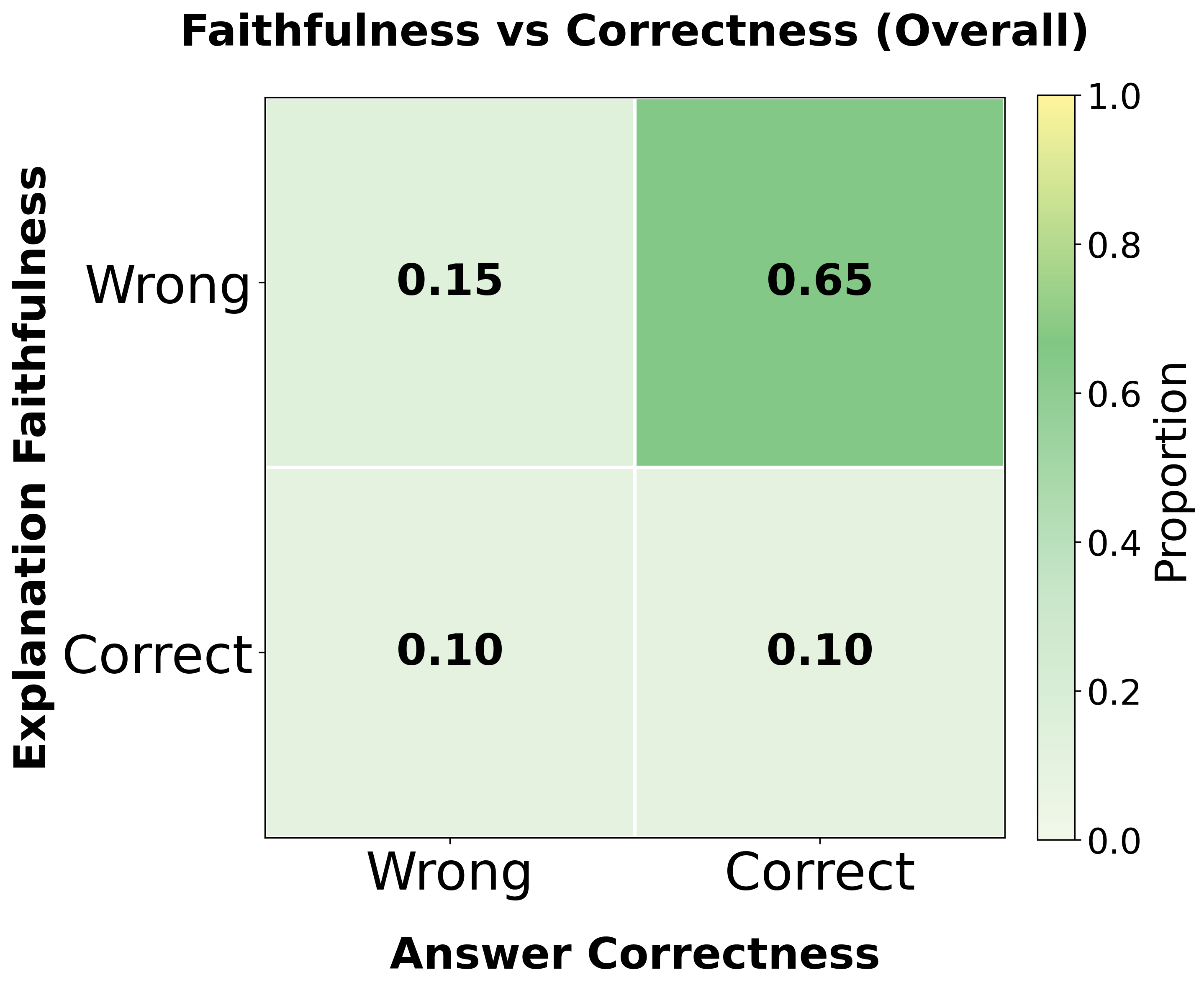}
        \caption{Reasoning faithfulness analysis.}
        \label{fig:hallucination}
    \end{subfigure}
    
    \caption{\textbf{Failure and faithfulness analysis of MLLM spatial reasoning.}
    (a) Human-annotated taxonomy of reasoning errors, where numeric and relational errors dominate.
    (b) Faithfulness matrix categorizing predictions into faithful reasoning, decision errors, lucky guesses, and hallucination.}
    \label{fig:failure_faithfulness}
    \vspace{-1mm}
\end{figure}

\textbf{Human validation.} To quantify the limitations of current MLLMs in spatial reasoning, we manually analyze and categorize model failures into four major types, as illustrated in Figure~\ref{fig:error_analysis}. These error categories reflect cognitive and perceptual challenges encountered by MLLMs in spatial reasoning over multimodal medical data: \textbf{(1) Visual perception error}, which arises when models fail to recognize or localize visual entities in medical images, such as misidentifying anatomical structures, missing target regions, or misinterpreting spatial layouts in different views; 
\textbf{(2) Linguistic intelligence error}, stems from the error in the language artifact, format output, and logic error in reasoning;
\textbf{(3) Relational reasoning error}, which failures in modeling spatial relationships among anatomical structures or lesions, such as incorrect reasoning about relative position, size, direction; and  \textbf{(4) Numeric reasoning error}, which occurs when models incorrectly perform quantitative reasoning, including volumes, and ratios derived from visual or textual information. Results from the SpatialMed suggest that: \textbf{\emph{The spatial reasoning, and the language to interpret the model's spatial understanding, is the challenge of the MLLMs}}.

\textbf{Reasoning Faithfulness.}
Human validation reveals frequent hallucination and random generation in MLLM spatial reasoning (Figure~\ref{fig:hallucination}). To disentangle reasoning quality from prediction accuracy, we assess whether the generated chain-of-thought actually supports the ground-truth answer, yielding four outcomes:
\emph{(i)} \textit{faithful reasoning} (explanation and answer both correct);
\emph{(ii)} \textit{decision errors} (correct explanation, wrong answer);
\emph{(iii)} \textit{lucky guesses} (wrong explanation, correct answer); and
\emph{(iv)} \textit{hallucinated reasoning} (both incorrect).
The results expose fundamental limitations in spatial understanding and hallucination, raising critical concerns for the safety and trustworthiness of MLLMs in medical imaging.

From the SpatialMed benchmark, we emphasize that: \textbf{\emph{It is necessary for the faithfulness-aware evaluation and training strategies in medical spatial reasoning.}}
\textbf{Model Stability and Numeric Commitment.}
Many evaluated models returned \emph{NaN} for absolute volume estimation, exposing a lack of \textbf{numeric commitment}: MLLMs prioritize qualitative recognition over calibrated measurement, underscoring the need for numeric supervision, unit-aware training, and structured output constraints for clinical use.

\label{sec:experiments}
\vspace{-4mm}
\section{Conclusion}
\vspace{-3mm}
In conclusion, we present SpatialMed, the first benchmark for evaluating 3D spatial reasoning in medical MLLMs, comprising 31,253 QA pairs across 2375 CT scans. Our evaluation of 24 state-of-the-art MLLMs and extensive analyses reveals substantial limitations: even top performers achieve only modest accuracy above a random baseline, with pronounced failures in relative position reasoning and numerical estimation, and the majority of reasoning chains are hallucinated. These findings demonstrate that current MLLMs lack the fundamentals for robust spatial reasoning in clinical applications. Future directions include spatial chain-of-thought reasoning, self-supervised spatial objectives, structured numeric output constraints, and faithfulness-aware training. We release the SpatialMed and the dataset creation pipeline to advance spatially-aware medical AI development.

\bibliographystyle{plain}
\bibliography{example_paper}

@article{llavamed,
  title={Llava-med: Training a large language-and-vision assistant for biomedicine in one day},
  author={Li, Chunyuan and Wong, Cliff and Zhang, Sheng and Usuyama, Naoto and Liu, Haotian and Yang, Jianwei and Naumann, Tristan and Poon, Hoifung and Gao, Jianfeng},
  journal={Advances in Neural Information Processing Systems},
  volume={36},
  pages={28541--28564},
  year={2023}
}

@article{bai2025qwen3,
  title={Qwen3-vl technical report},
  author={Bai, Shuai and Cai, Yuxuan and Chen, Ruizhe and Chen, Keqin and Chen, Xionghui and Cheng, Zesen and Deng, Lianghao and Ding, Wei and Gao, Chang and Ge, Chunjiang and others},
  journal={arXiv preprint arXiv:2511.21631},
  year={2025}
}

@article{singh2025openai,
  title={Openai gpt-5 system card},
  author={Singh, Aaditya and Fry, Adam and Perelman, Adam and Tart, Adam and Ganesh, Adi and El-Kishky, Ahmed and McLaughlin, Aidan and Low, Aiden and Ostrow, AJ and Ananthram, Akhila and others},
  journal={arXiv preprint arXiv:2601.03267},
  year={2025}
}

@article{xin2025med3dvlm,
  title={Med3DVLM: An Efficient Vision-Language Model for 3D Medical Image Analysis},
  author={Xin, Yu and Ates, Gorkem Can and Gong, Kuang and Shao, Wei},
  journal={IEEE Journal of Biomedical and Health Informatics},
  year={2025}
}

@article{comanici2025gemini,
  title={Gemini 2.5: Pushing the frontier with advanced reasoning, multimodality, long context, and next generation agentic capabilities},
  author={Comanici, Gheorghe and Bieber, Eric and Schaekermann, Mike and Pasupat, Ice and Sachdeva, Noveen and Dhillon, Inderjit and Blistein, Marcel and Ram, Ori and Zhang, Dan and Rosen, Evan and others},
  journal={arXiv preprint arXiv:2507.06261},
  year={2025}
}

@inproceedings{hamamcibetter,
  title={Better Tokens for Better 3D: Advancing Vision-Language Modeling in 3D Medical Imaging},
  author={Hamamci, Ibrahim Ethem and Er, Sezgin and Shit, Suprosanna and Reynaud, Hadrien and Yang, Dong and Guo, Pengfei and Edgar, Marc and Xu, Daguang and Kainz, Bernhard and Menze, Bjoern},
  booktitle={The Thirty-ninth Annual Conference on Neural Information Processing Systems}
}

@misc{qwen35,
    title  = {{Qwen3.5}: Towards Native Multimodal Agents},
    author = {{Qwen Team}},
    month  = {February},
    year   = {2026},
    url    = {https://qwen.ai/blog?id=qwen3.5}
}

@misc{lin2025healthgptmedicallargevisionlanguage,
      title={HealthGPT: A Medical Large Vision-Language Model for Unifying Comprehension and Generation via Heterogeneous Knowledge Adaptation}, 
      author={Tianwei Lin and Wenqiao Zhang and Sijing Li and Yuqian Yuan and Binhe Yu and Haoyuan Li and Wanggui He and Hao Jiang and Mengze Li and Xiaohui Song and Siliang Tang and Jun Xiao and Hui Lin and Yueting Zhuang and Beng Chin Ooi},
      year={2025},
      eprint={2502.09838},
      archivePrefix={arXiv},
      primaryClass={cs.CV},
      url={https://arxiv.org/abs/2502.09838}, 
}

@article{internvl35,
  title={InternVL3.5: Advancing Open-Source Multimodal Models in Versatility, Reasoning, and Efficiency},
  author={Wang, Weiyun and Gao, Zhangwei and Gu, Lixin and Pu, Hengjun and Cui, Long and Wei, Xingguang and Liu, Zhaoyang and Jing, Linglin and Ye, Shenglong and Shao, Jie and others},
  journal={arXiv preprint arXiv:2508.18265},
  year={2025}
}

@article{unimedvl,
  title={Unimedvl: Unifying medical multimodal understanding and generation through observation-knowledge-analysis},
  author={Ning, Junzhi and Li, Wei and Tang, Cheng and Lin, Jiashi and Ma, Chenglong and Zhang, Chaoyang and Liu, Jiyao and Chen, Ying and Gao, Shujian and Liu, Lihao and others},
  journal={arXiv preprint arXiv:2510.15710},
  year={2025}
}

@inproceedings{medflamingo,
  title={Med-flamingo: a multimodal medical few-shot learner},
  author={Moor, Michael and Huang, Qian and Wu, Shirley and Yasunaga, Michihiro and Dalmia, Yash and Leskovec, Jure and Zakka, Cyril and Reis, Eduardo Pontes and Rajpurkar, Pranav},
  booktitle={Machine Learning for Health (ML4H)},
  pages={353--367},
  year={2023},
  organization={PMLR}
}

@article{zeng2025glm,
  title={Glm-4.5: Agentic, reasoning, and coding (arc) foundation models},
  author={Zeng, Aohan and Lv, Xin and Zheng, Qinkai and Hou, Zhenyu and Chen, Bin and Xie, Chengxing and Wang, Cunxiang and Yin, Da and Zeng, Hao and Zhang, Jiajie and others},
  journal={arXiv preprint arXiv:2508.06471},
  year={2025}
}

@article{sammed2d,
  title={Sa-med2d-20m dataset: Segment anything in 2d medical imaging with 20 million masks},
  author={Ye, Jin and Cheng, Junlong and Chen, Jianpin and Deng, Zhongying and Li, Tianbin and Wang, Haoyu and Su, Yanzhou and Huang, Ziyan and Chen, Jilong and Jiang, Lei and others},
  journal={arXiv preprint arXiv:2311.11969},
  year={2023}
}

@article{medmoe,
  title={Med-moe: Mixture of domain-specific experts for lightweight medical vision-language models},
  author={Jiang, Songtao and Zheng, Tuo and Zhang, Yan and Jin, Yeying and Yuan, Li and Liu, Zuozhu},
  journal={arXiv preprint arXiv:2404.10237},
  year={2024}
}

@inproceedings{internvl,
  title={Internvl: Scaling up vision foundation models and aligning for generic visual-linguistic tasks},
  author={Chen, Zhe and Wu, Jiannan and Wang, Wenhai and Su, Weijie and Chen, Guo and Xing, Sen and Zhong, Muyan and Zhang, Qinglong and Zhu, Xizhou and Lu, Lewei and others},
  booktitle={Proceedings of the IEEE/CVF conference on computer vision and pattern recognition},
  pages={24185--24198},
  year={2024}
}

@article{medvlm-r1,
  title={Medvlm-r1: Incentivizing medical reasoning capability of vision-language models (vlms) via reinforcement learning},
  author={Pan, Jiazhen and Liu, Che and Wu, Junde and Liu, Fenglin and Zhu, Jiayuan and Li, Hongwei Bran and Chen, Chen and Ouyang, Cheng and Rueckert, Daniel},
  journal={arXiv preprint arXiv:2502.19634},
  year={2025}
}

@misc{liu2024llavanext,
    title={LLaVA-NeXT: Improved reasoning, OCR, and world knowledge},
    url={https://llava-vl.github.io/blog/2024-01-30-llava-next/},
    author={Liu, Haotian and Li, Chunyuan and Li, Yuheng and Li, Bo and Zhang, Yuanhan and Shen, Sheng and Lee, Yong Jae},
    month={January},
    year={2024}
}

@article{chen2024huatuogpt,
  title={Huatuogpt-vision, towards injecting medical visual knowledge into multimodal llms at scale},
  author={Chen, Junying and Gui, Chi and Ouyang, Ruyi and Gao, Anningzhe and Chen, Shunian and Chen, Guiming Hardy and Wang, Xidong and Zhang, Ruifei and Cai, Zhenyang and Ji, Ke and others},
  journal={arXiv preprint arXiv:2406.19280},
  year={2024}
}

@misc{cheng2023sammed2d,
      title={SAM-Med2D}, 
      author={Junlong Cheng and Jin Ye and Zhongying Deng and Jianpin Chen and Tianbin Li and Haoyu Wang and Yanzhou Su and Ziyan Huang and Jilong Chen and Lei Jiangand Hui Sun and Junjun He and Shaoting Zhang and Min Zhu and Yu Qiao},
      year={2023},
      eprint={2308.16184},
      archivePrefix={arXiv},
      primaryClass={cs.CV}
}

@article{qwen2vl,
  title={Qwen2-vl: Enhancing vision-language model's perception of the world at any resolution},
  author={Wang, Peng and Bai, Shuai and Tan, Sinan and Wang, Shijie and Fan, Zhihao and Bai, Jinze and Chen, Keqin and Liu, Xuejing and Wang, Jialin and Ge, Wenbin and others},
  journal={arXiv preprint arXiv:2409.12191},
  year={2024}
}

@article{llama3,
  title={The llama 3 herd of models},
  author={Grattafiori, Aaron and Dubey, Abhimanyu and Jauhri, Abhinav and Pandey, Abhinav and Kadian, Abhishek and Al-Dahle, Ahmad and Letman, Aiesha and Mathur, Akhil and Schelten, Alan and Vaughan, Alex and others},
  journal={arXiv preprint arXiv:2407.21783},
  year={2024}
}

@misc{ctrate1,
      title={Developing Generalist Foundation Models from a Multimodal Dataset for 3D Computed Tomography}, 
      author={Ibrahim Ethem Hamamci and Sezgin Er and Furkan Almas and Ayse Gulnihan Simsek and Sevval Nil Esirgun and Irem Dogan and Muhammed Furkan Dasdelen and Omer Faruk Durugol and Bastian Wittmann and Tamaz Amiranashvili and Enis Simsar and Mehmet Simsar and Emine Bensu Erdemir and Abdullah Alanbay and Anjany Sekuboyina and Berkan Lafci and Christian Bluethgen and Mehmet Kemal Ozdemir and Bjoern Menze},
      year={2024},
      eprint={2403.17834},
      archivePrefix={arXiv},
      primaryClass={cs.CV},
      url={https://arxiv.org/abs/2403.17834}, 
}

@inproceedings{ctrate2,
      title={Generatect: Text-conditional generation of 3d chest ct volumes},
      author={Hamamci, Ibrahim Ethem and Er, Sezgin and Sekuboyina, Anjany and Simsar, Enis and Tezcan, Alperen and Simsek, Ayse Gulnihan and Esirgun, Sevval Nil and Almas, Furkan and Do{\u{g}}an, Irem and Dasdelen, Muhammed Furkan and others},
      booktitle={European Conference on Computer Vision},
      pages={126--143},
      year={2024},
      organization={Springer}
}

@article{bai2024m3d,
  title={M3d: Advancing 3d medical image analysis with multi-modal large language models},
  author={Bai, Fan and Du, Yuxin and Huang, Tiejun and Meng, Max Q-H and Zhao, Bo},
  journal={arXiv preprint arXiv:2404.00578},
  year={2024}
}

@inproceedings{bassi2025radgpt,
  title={Radgpt: Constructing 3d image-text tumor datasets},
  author={Bassi, Pedro RAS and Yavuz, Mehmet Can and Hamamci, Ibrahim Ethem and Er, Sezgin and Chen, Xiaoxi and Li, Wenxuan and Menze, Bjoern and Decherchi, Sergio and Cavalli, Andrea and Wang, Kang and others},
  booktitle={Proceedings of the IEEE/CVF International Conference on Computer Vision},
  pages={23720--23730},
  year={2025}
}

@article{ctchat,
  title={A foundation model utilizing chest ct volumes and radiology reports for supervised-level zero-shot detection of abnormalities},
  author={Hamamci, Ibrahim Ethem and Er, Sezgin and Almas, Furkan and Simsek, Ayse Gulnihan and Esirgun, Sevval Nil and Dogan, Irem and Dasdelen, Muhammed Furkan and Wittmann, Bastian and Simsar, Enis and Simsar, Mehmet and others},
  journal={CoRR},
  year={2024}
}

@article{brats1,
  title={The multimodal brain tumor image segmentation benchmark (BRATS)},
  author={Menze, Bjoern H and Jakab, Andras and Bauer, Stefan and Kalpathy-Cramer, Jayashree and Farahani, Keyvan and Kirby, Justin and Burren, Yuliya and Porz, Nicole and Slotboom, Johannes and Wiest, Roland and others},
  journal={IEEE transactions on medical imaging},
  volume={34},
  number={10},
  pages={1993--2024},
  year={2014},
  publisher={IEEE}
}

@article{brats2,
  title={Advancing the cancer genome atlas glioma MRI collections with expert segmentation labels and radiomic features},
  author={Bakas, Spyridon and Akbari, Hamed and Sotiras, Aristeidis and Bilello, Michel and Rozycki, Martin and Kirby, Justin S and Freymann, John B and Farahani, Keyvan and Davatzikos, Christos},
  journal={Scientific data},
  volume={4},
  number={1},
  pages={1--13},
  year={2017},
  publisher={Nature Publishing Group}
}

@article{brats3,
  title={Identifying the best machine learning algorithms for brain tumor segmentation, progression assessment, and overall survival prediction in the BRATS challenge},
  author={Bakas, Spyridon and Reyes, Mauricio and Jakab, Andras and Bauer, Stefan and Rempfler, Markus and Crimi, Alessandro and Shinohara, Russell Takeshi and Berger, Christoph and Ha, Sung Min and Rozycki, Martin and others},
  journal={arXiv preprint arXiv:1811.02629},
  year={2018}
}

@misc{kits,
      title={The KiTS21 Challenge: Automatic segmentation of kidneys, renal tumors, and renal cysts in corticomedullary-phase CT}, 
      author={Nicholas Heller and Fabian Isensee and Dasha Trofimova and Resha Tejpaul and Zhongchen Zhao and Huai Chen and Lisheng Wang and Alex Golts and Daniel Khapun and Daniel Shats and Yoel Shoshan and Flora Gilboa-Solomon and Yasmeen George and Xi Yang and Jianpeng Zhang and Jing Zhang and Yong Xia and Mengran Wu and Zhiyang Liu and Ed Walczak and Sean McSweeney and Ranveer Vasdev and Chris Hornung and Rafat Solaiman and Jamee Schoephoerster and Bailey Abernathy and David Wu and Safa Abdulkadir and Ben Byun and Justice Spriggs and Griffin Struyk and Alexandra Austin and Ben Simpson and Michael Hagstrom and Sierra Virnig and John French and Nitin Venkatesh and Sarah Chan and Keenan Moore and Anna Jacobsen and Susan Austin and Mark Austin and Subodh Regmi and Nikolaos Papanikolopoulos and Christopher Weight},
      year={2023},
      eprint={2307.01984},
      archivePrefix={arXiv},
      primaryClass={cs.CV}
}

@article{baharoon2025rexgroundingct,
  title={Rexgroundingct: A 3d chest ct dataset for segmentation of findings from free-text reports},
  author={Baharoon, Mohammed and Luo, Luyang and Moritz, Michael and Kumar, Abhinav and Kim, Sung Eun and Zhang, Xiaoman and Zhu, Miao and Alabbad, Mahmoud Hussain and Alhazmi, Maha Sbayel and Mistry, Neel P and others},
  journal={arXiv preprint arXiv:2507.22030},
  year={2025}
}

@article{gai20253d,
  title={3D-RAD: A Comprehensive 3D Radiology Med-VQA Dataset with Multi-Temporal Analysis and Diverse Diagnostic Tasks},
  author={Gai, Xiaotang and Liu, Jiaxiang and Li, Yichen and Meng, Zijie and Wu, Jian and Liu, Zuozhu},
  journal={arXiv preprint arXiv:2506.11147},
  year={2025}
}

@article{huang2023inspect,
  title={INSPECT: A Multimodal Dataset for Pulmonary Embolism Diagnosis and Prognosis},
  author={Huang, Shih-Cheng and Huo, Zepeng and Steinberg, Ethan and Chiang, Chia-Chun and Langlotz, Curtis and Lungren, Matthew P and Yeung, Serena and Shah, Nigam and Fries, Jason Alan},
  journal={arXiv preprint arXiv:2311.10798},
  year={2023}
}

@inproceedings{hamamci2024ct2rep,
      title={Ct2rep: Automated radiology report generation for 3d medical imaging},
      author={Hamamci, Ibrahim Ethem and Er, Sezgin and Menze, Bjoern},
      booktitle={International Conference on Medical Image Computing and Computer-Assisted Intervention},
      pages={476--486},
      year={2024},
      organization={Springer}
}

@misc{qwen3,
      title={Qwen3 Technical Report}, 
      author={Qwen},
      year={2025},
      eprint={2505.09388},
      archivePrefix={arXiv},
      primaryClass={cs.CL},
      url={https://arxiv.org/abs/2505.09388}, 
}

@article{glm,
  title={GLM-4.1 V-Thinking: Towards Versatile Multimodal Reasoning with Scalable Reinforcement Learning},
  author={Hong, Wenyi and Yu, Wenmeng and Gu, Xiaotao and Wang, Guo and Gan, Guobing and Tang, Haomiao and Cheng, Jiale and Qi, Ji and Ji, Junhui and Pan, Lihang and others},
  journal={arXiv preprint arXiv:2507.01006},
  year={2025}
}

@inproceedings{chen2024medblip,
  title={Medblip: Bootstrapping language-image pre-training from 3d medical images and texts},
  author={Chen, Qiuhui and Hong, Yi},
  booktitle={Proceedings of the Asian conference on computer vision},
  pages={2404--2420},
  year={2024}
}

@article{shi2024med,
  title={Med-2e3: A 2d-enhanced 3d medical multimodal large language model},
  author={Shi, Yiming and Zhu, Xun and Wang, Kaiwen and Hu, Ying and Guo, Chenyi and Li, Miao and Wu, Ji},
  journal={arXiv preprint arXiv:2411.12783},
  year={2024}
}

@inproceedings{gqa,
  title={Gqa: A new dataset for real-world visual reasoning and compositional question answering},
  author={Hudson, Drew A and Manning, Christopher D},
  booktitle={Proceedings of the IEEE/CVF conference on computer vision and pattern recognition},
  pages={6700--6709},
  year={2019}
}

@article{gsr,
  title={Gsr-bench: A benchmark for grounded spatial reasoning evaluation via multimodal llms},
  author={Rajabi, Navid and Kosecka, Jana},
  journal={arXiv preprint arXiv:2406.13246},
  year={2024}
}

@inproceedings{vissi,
  title={Thinking in space: How multimodal large language models see, remember, and recall spaces},
  author={Yang, Jihan and Yang, Shusheng and Gupta, Anjali W and Han, Rilyn and Fei-Fei, Li and Xie, Saining},
  booktitle={Proceedings of the Computer Vision and Pattern Recognition Conference},
  pages={10632--10643},
  year={2025}
}

@article{yi2019clevrer,
  title={Clevrer: Collision events for video representation and reasoning},
  author={Yi, Kexin and Gan, Chuang and Li, Yunzhu and Kohli, Pushmeet and Wu, Jiajun and Torralba, Antonio and Tenenbaum, Joshua B},
  journal={arXiv preprint arXiv:1910.01442},
  year={2019}
}

@article{trinh2025prs,
  title={Prs-med: Position reasoning segmentation with vision-language model in medical imaging},
  author={Trinh, Quoc-Huy and Nguyen, Minh-Van and Zeng, Jung and Bagci, Ulas and Jha, Debesh},
  journal={arXiv preprint arXiv:2505.11872},
  year={2025}
}

@article{canese2013pubmed,
  title={PubMed: the bibliographic database},
  author={Canese, Kathi and Weis, Sarah},
  journal={The NCBI handbook},
  volume={2},
  number={1},
  pages={2013},
  year={2013},
  publisher={National Center for Biotechnology Information (US) Bethesda, MD}
}

@article{ji2022amos,
  title={Amos: A large-scale abdominal multi-organ benchmark for versatile medical image segmentation},
  author={Ji, Yuanfeng and Bai, Haotian and Ge, Chongjian and Yang, Jie and Zhu, Ye and Zhang, Ruimao and Li, Zhen and Zhanng, Lingyan and Ma, Wanling and Wan, Xiang and others},
  journal={Advances in neural information processing systems},
  volume={35},
  pages={36722--36732},
  year={2022}
}

@article{wasserthal2023totalsegmentator,
  title={TotalSegmentator: robust segmentation of 104 anatomic structures in CT images},
  author={Wasserthal, Jakob and Breit, Hanns-Christian and Meyer, Manfred T and Pradella, Maurice and Hinck, Daniel and Sauter, Alexander W and Heye, Tobias and Boll, Daniel T and Cyriac, Joshy and Yang, Shan and others},
  journal={Radiology: Artificial Intelligence},
  volume={5},
  number={5},
  pages={e230024},
  year={2023},
  publisher={Radiological Society of North America}
}

@article{kits1,
  title={The state of the art in kidney and kidney tumor segmentation in contrast-enhanced CT imaging: Results of the KiTS19 Challenge},
  author={Heller, Nicholas and Isensee, Fabian and Maier-Hein, Klaus H and Hou, Xiaoshuai and Xie, Chunmei and Li, Fengyi and Nan, Yang and Mu, Guangrui and Lin, Zhiyong and Han, Miofei and others},
  journal={Medical Image Analysis},
  pages={101821},
  year={2020},
  publisher={Elsevier}
}

@article{kits2,
  title={The kits19 challenge data: 300 kidney tumor cases with clinical context, ct semantic segmentations, and surgical outcomes},
  author={Heller, Nicholas and Sathianathen, Niranjan and Kalapara, Arveen and Walczak, Edward and Moore, Keenan and Kaluzniak, Heather and Rosenberg, Joel and Blake, Paul and Rengel, Zachary and Oestreich, Makinna and others},
  journal={arXiv preprint arXiv:1904.00445},
  year={2019}
}

@article{antonelli2022medical,
  title={The medical segmentation decathlon},
  author={Antonelli, Michela and Reinke, Annika and Bakas, Spyridon and Farahani, Keyvan and Kopp-Schneider, Annette and Landman, Bennett A and Litjens, Geert and Menze, Bjoern and Ronneberger, Olaf and Summers, Ronald M and others},
  journal={Nature communications},
  volume={13},
  number={1},
  pages={4128},
  year={2022},
  publisher={Nature Publishing Group UK London}
}

@article{lewis2020retrieval,
  title={Retrieval-augmented generation for knowledge-intensive nlp tasks},
  author={Lewis, Patrick and Perez, Ethan and Piktus, Aleksandra and Petroni, Fabio and Karpukhin, Vladimir and Goyal, Naman and K{\"u}ttler, Heinrich and Lewis, Mike and Yih, Wen-tau and Rockt{\"a}schel, Tim and others},
  journal={Advances in neural information processing systems},
  volume={33},
  pages={9459--9474},
  year={2020}
}

@article{hamamci2024developing,
  title={Developing generalist foundation models from a multimodal dataset for 3d computed tomography},
  author={Hamamci, Ibrahim Ethem and Er, Sezgin and Wang, Chenyu and Almas, Furkan and Simsek, Ayse Gulnihan and Esirgun, Sevval Nil and Dogan, Irem and Durugol, Omer Faruk and Hou, Benjamin and Shit, Suprosanna and others},
  journal={arXiv preprint arXiv:2403.17834},
  year={2024}
}

@inproceedings{chen2024bimcv,
  title={Bimcv-r: A landmark dataset for 3d ct text-image retrieval},
  author={Chen, Yinda and Liu, Che and Liu, Xiaoyu and Arcucci, Rossella and Xiong, Zhiwei},
  booktitle={International Conference on Medical Image Computing and Computer-Assisted Intervention},
  pages={124--134},
  year={2024},
  organization={Springer}
}

@article{zhu2025internvl3,
  title={Internvl3: Exploring advanced training and test-time recipes for open-source multimodal models},
  author={Zhu, Jinguo and Wang, Weiyun and Chen, Zhe and Liu, Zhaoyang and Ye, Shenglong and Gu, Lixin and Tian, Hao and Duan, Yuchen and Su, Weijie and Shao, Jie and others},
  journal={arXiv preprint arXiv:2504.10479},
  year={2025}
}

@article{butsanets2025radimagenet,
  title={RadImageNet-VQA: A Large-Scale CT and MRI Dataset for Radiologic Visual Question Answering},
  author={Butsanets, L{\'e}o and Corbi{\`e}re, Charles and Khlaut, Julien and Manceron, Pierre and Dancette, Corentin},
  journal={arXiv preprint arXiv:2512.17396},
  year={2025}
}

@article{cai2024internlm2,
  title={Internlm2 technical report},
  author={Cai, Zheng and Cao, Maosong and Chen, Haojiong and Chen, Kai and Chen, Keyu and Chen, Xin and Chen, Xun and Chen, Zehui and Chen, Zhi and Chu, Pei and others},
  journal={arXiv preprint arXiv:2403.17297},
  year={2024}
}

@inproceedings{ribeiro2016should,
  title={" Why should i trust you?" Explaining the predictions of any classifier},
  author={Ribeiro, Marco Tulio and Singh, Sameer and Guestrin, Carlos},
  booktitle={Proceedings of the 22nd ACM SIGKDD international conference on knowledge discovery and data mining},
  pages={1135--1144},
  year={2016}
}

@article{sellergren2025medgemma,
  title={Medgemma technical report},
  author={Sellergren, Andrew and Kazemzadeh, Sahar and Jaroensri, Tiam and Kiraly, Atilla and Traverse, Madeleine and Kohlberger, Timo and Xu, Shawn and Jamil, Fayaz and Hughes, C{\'\i}an and Lau, Charles and others},
  journal={arXiv preprint arXiv:2507.05201},
  year={2025}
}

@article{qwen3embedding,
  title={Qwen3 Embedding: Advancing Text Embedding and Reranking Through Foundation Models},
  author={Zhang, Yanzhao and Li, Mingxin and Long, Dingkun and Zhang, Xin and Lin, Huan and Yang, Baosong and Xie, Pengjun and Yang, An and Liu, Dayiheng and Lin, Junyang and Huang, Fei and Zhou, Jingren},
  journal={arXiv preprint arXiv:2506.05176},
  year={2025}
}

@article{johnson2019billion,
  title={Billion-scale similarity search with {GPUs}},
  author={Johnson, Jeff and Douze, Matthijs and J{\'e}gou, Herv{\'e}},
  journal={IEEE Transactions on Big Data},
  volume={7},
  number={3},
  pages={535--547},
  year={2019},
  publisher={IEEE}
}

@article{shen2024thermometer,
  title={Thermometer: Towards universal calibration for large language models},
  author={Shen, Maohao and Das, Subhro and Greenewald, Kristjan and Sattigeri, Prasanna and Wornell, Gregory and Ghosh, Soumya},
  journal={arXiv preprint arXiv:2403.08819},
  year={2024}
}
\newpage
\appendix



\section{Appendix}

\subsection{Broader Impact}

This paper introduces a new benchmark and data generation pipeline to study spatial reasoning in medical vision language models using 3D CT scans. The main goal of this work is to advance machine learning methods for understanding spatial relations in medical images, which is an important step toward more reliable AI systems in healthcare.

\textbf{Potential positive impact.}
Improved spatial reasoning in medical AI systems may support clinical practice in several ways. Models that better understand the comparison, volume, and relative position of anatomical structures could assist with image interpretation, treatment planning, and medical education. This may help reduce workload for clinicians and improve consistency in image-based analysis. The proposed benchmark may also encourage more transparent evaluation of model limitations, which is important for safety in medical AI.

\textbf{Risks and limitations.}
Despite these potential benefits, there are risks. First, models evaluated on this benchmark may still make errors in real clinical settings. Good performance on a benchmark does not guarantee safe or correct behavior in practice. Second, there is a risk that users may over trust model outputs, especially for quantitative values such as volume or bounding boxes extraction. Such errors could lead to incorrect clinical decisions if systems are used without proper human oversight. The dataset is built from existing medical imaging data with segmentation masks. Although the data are processed for research use, any use of medical data raises concerns about privacy and data governance. We follow the licenses and usage rules of the source datasets. Future work should continue to consider data protection and responsible sharing.

\textbf{Future societal considerations.}
As medical vision language models become more capable, they may be integrated into clinical tools. This increases the need for clear evaluation standards, human-in-the-loop design, and regulatory review. Our benchmark highlights current weaknesses in spatial and numeric reasoning, which may help prevent premature deployment of unreliable systems.

\subsection{Details of LLM usage}
In our case, we use LLMs as the core of our agent system to support the dataset generation. Furthermore, the Multimodal Large Language Models across domains are used to verify the ability of spatial understanding.
\subsection{Prompt for Dataset Creation}
\label{section:appen_prompt}
In this section, we provide the additional details of the prompt for Spatial QA Generation, Filtering Process, and Trivial Validation as follows:

\begin{tcolorbox}[
    title=Prompt for Spatial QA Generation,
    colback=gray!5,
    colframe=black,
    fonttitle=\bfseries,
    boxrule=0.5pt
]
\small
\emph{{[Medical context retrieved from the RAG pipeline]}}
You are a radiologist analyzing a 3D medical volume. You are given:
\begin{itemize}
    \item {[a list of anatomical volumes]},
    \item {[a set of 3D bounding boxes]}.
\end{itemize}

Example question--answer pairs are provided below:
\begin{quote}
\emph{[List of example question--answer pairs]}
\end{quote}

\noindent
\textbf{Instruction.} Based on the provided spatial information, and your knowledge for the \emph{tumor names}, generate 10 question--answer pairs that require clinically meaningful spatial reasoning. 
The questions and answers must be medically relevant and consistent with a radiologist’s viewpoint, and must be diverse in the medical context linguistic. Return response in tempate \emph{Question: {[Your question here]} - Answer: {[Your answer here]}}
\end{tcolorbox}

\begin{tcolorbox}[
    title=Prompt for quality filtering,
    colback=gray!5,
    colframe=black,
    fonttitle=\bfseries,
    boxrule=0.5pt
]
\small
You are a radiologist, and you are seeing the 3D volumes. You are given:
Context: \emph{Relevant contexts to the question and answer pair}
\begin{itemize}
    \item {[a list of anatomical volumes]},
    \item {[a set of 3D bounding boxes]}.
\end{itemize}

You are giving a question and an answer pair: \emph{Question and answer pair}

\noindent
\textbf{Instruction.} From your knowledge, evaluate if this question and answer pairs meet these criteria: (1) contain medical content, (2) the correctness of the content with the context you know.

Return only 0 for unqualified, and 1 for qualified data. 
\end{tcolorbox}

\begin{tcolorbox}[
    title=Prompt for trivial validation,
    colback=gray!5,
    colframe=black,
    fonttitle=\bfseries,
    boxrule=0.5pt
]
\small
\textbf{Context:} \emph{Medical Context relevant to question retrieved from RAG pipeline} \\
\noindent You are a doctor, and you get these question and answers from the client. \\
\noindent \textbf{Instruction:} With the provided relevant context you are provided, answer the question:
\begin{quote}
\emph{Question and choices}
\end{quote}

\noindent Return the final answer only, for example, "A. liver.", do not return anything else.

\end{tcolorbox}

For the spatial annotations input for the prompt, they are calculated by the tools, mentioned in Section~\ref{subsec:datasource}.

During the generation process, by using few-shot prompting and provide the context from the RAG pipeline, the agents are provided the medical knowledge, therefore enhance the capability to generate robust medical content robust, and ensure the linguistic quality of the question-answer samples. Moreover, all of stages, we also leverage the instruction prompting template for the output, which can help us to ensure the correct format of the generation from the models.

\subsection{Prompt for Benchmark}
\label{section:appen_benchmark_prompt}
In this section, we provide the prompt templates used to query the multimodal large language models during benchmarking. Since our benchmark covers both 2D multi-view vision-language models and 3D volume-based medical language models, we design two corresponding prompt templates to accommodate their different input modalities. In both cases, we enforce a strict output format to ensure reliable answer parsing and fair comparison across models.

\begin{tcolorbox}[
    title=Prompt for 2D Multi-View Benchmark,
    colback=gray!5,
    colframe=black,
    fonttitle=\bfseries,
    boxrule=0.5pt
]
\small
\textbf{System Prompt:} \\
You are an expert medical imaging assistant. Answer the user's multiple-choice questions based ONLY on the provided images. Return the final answer only.
\begin{itemize}
    \item If the options are A/B/C/D, respond exactly like: \emph{Answer: A}
\end{itemize}
Do not include explanations. \\

\noindent \textbf{User Input:} \\
\emph{[Axial view image]}, \emph{[Sagittal view image]}, \emph{[Coronal view image]} \\
\emph{[Question and answer choices]}
\end{tcolorbox}

For 2D multimodal models, we provide three orthogonal views (axial, sagittal, and coronal) simultaneously, allowing the models to leverage complementary spatial cues across all three dimensions, consistent with how radiologists interpret volumetric CT data in clinical practice.

\begin{tcolorbox}[
    title=Prompt for 3D Volume Benchmark,
    colback=gray!5,
    colframe=black,
    fonttitle=\bfseries,
    boxrule=0.5pt
]
\small
\textbf{System Prompt:} \\
You are a medical QA assistant. Return ONLY a numeric volume followed by \emph{cm\textsuperscript{3}}. Example: \emph{12.5 cm\textsuperscript{3}}. \\

\noindent \textbf{User Input:} \\
Anatomy: \emph{[Target anatomical structure]} \\
Question: \emph{[Spatial reasoning question]}
\end{tcolorbox}

For 3D volume-based medical language models, the input is the full volumetric scan paired with metadata describing the case and the target anatomy. The output is constrained to a numeric value with an explicit unit (\emph{cm\textsuperscript{3}}), which enables automatic extraction of predicted volumes via regular expression matching and allows direct computation of absolute error against the ground-truth measurements.

By enforcing strict output templates in both settings, we minimize ambiguity in answer extraction and ensure that the evaluation reflects the model's spatial reasoning capability rather than its output formatting behavior.

\subsection{Retrieved Augmented Generation Details}

In terms of the Retrieved Augmented Generation pipeline, we leverage the Qwen3-8B-Embedding model~\cite{qwen3embedding} as the based embedding model to extract all embeddings from the text of the PubMed~\cite{canese2013pubmed} dataset. All of the embeddings are stored via the index storage of Faiss~\cite {johnson2019billion}. During the inference process of the agent, the top 5 most relevant records are chosen as the input context for the agent.

\subsection{Answer Parsing Rules}
\label{section:appen_parsing}
Since the outputs of large language models can vary in format despite explicit instruction prompting, we design deterministic parsing rules to extract the final answer from the raw model output. We adopt two parsing strategies corresponding to the two benchmark settings: multiple-choice parsing for the 2D multi-view benchmark, and numeric volume parsing for the 3D volume benchmark.

\paragraph{Multiple-Choice Parsing.}
For the 2D multi-view benchmark, we apply a cascaded regular expression matching strategy to the raw model output. The parser first searches for the explicit template \emph{``Answer: X''}, where \emph{X} is a letter from \{A, B, C, D\}. If this pattern is not found, the parser falls back to searching for any standalone occurrence of a valid option letter in the output. The matching is case-insensitive, and the extracted answer is normalized to uppercase. Ground-truth answers are normalized using the same procedure prior to comparison, ensuring consistent evaluation across models with different output styles. A prediction is considered correct only if the normalized prediction exactly matches the normalized ground truth.

\paragraph{Numeric Volume Parsing.}
For the 3D volume benchmark, we extract the predicted volume from the model output using a regular expression that matches a floating-point number immediately followed by the unit \emph{cm\textsuperscript{3}} (accepting both the Unicode superscript and the plain \emph{cm3} notation). The first matched value is taken as the predicted volume. If no valid pattern is found, the prediction is marked as unparseable and excluded from the error computation. For successfully parsed predictions, we compute the Mean Relative Accuracy (mentioned in Equation~\ref{eqa:mra} between the predicted value and the ground-truth volume, then we report the MRA over all parsed samples as the primary evaluation metric.

By applying these deterministic parsing rules, we ensure that the reported performance reflects the model's reasoning capability rather than artifacts of output formatting variability, and that the evaluation remains reproducible across different runs and model families.

\textbf{Parsing success rate.} Table~\ref{tab:parse_success} reports parse success on both the MCQ (spatial reasoning) and Numerical (volume estimation) tasks. Most modern general-domain 2D models (Qwen3-VL 4B/8B, InternVL3.5 14B/30B-A3B) achieve 100\% on both formats, and the top 3D model Med3DVLM parses reliably on MCQ (98.53\%) though it drops on numerical outputs (70.21\%), indicating that the parsing rule is not the bottleneck for these systems. Several models show markedly lower rates on at least one format: GLM-4.1V-9B (50.25\% MCQ) and GLM-4.6V-Flash (61.06\% MCQ, 10.91\% Num) frequently deviate from the requested formats, while Qwen3.5 9B degrades sharply on numerical outputs (13.40\%). The medical pretraining block is the most affected—HuatuoGPT-Vision (15.97\% MCQ), Med-VLM R1 (18.60\% MCQ), and MedMoE (47.52\% MCQ) fail to produce parseable answers on most MCQ questions, and MedMoE, Med-VLM R1, HuatuoGPT-Vision, and M3D all collapse to 0\% on numerical outputs, which demonstrates that these methods need improvement in the model architecture and training recipe to enable them to do the question answering in the quantitative tasks, which is the future work we expect to do.

\begin{table*}[!ht]
\centering
\caption{\textbf{Answer parse-success rate across evaluated models.} Rates are split by task format: MCQ (multiple-choice questions on the spatial reasoning benchmark) and Numerical (volume-estimation benchmark). Low rates indicate frequent format deviations that confound accuracy comparisons. The highest and lowest rates in each column are highlighted in \textbf{bold} and \textcolor{red}{red}, respectively.}
\label{tab:parse_success}
\setlength{\tabcolsep}{4pt}
\renewcommand{\arraystretch}{1.1}
\small
\begin{tabular}{@{}lcc@{\hspace{1.4em}}lcc@{\hspace{1.4em}}lcc@{}}
\toprule
\multicolumn{3}{c}{\textbf{2D Non-medical}} &
\multicolumn{3}{c}{\textbf{2D Medical}} &
\multicolumn{3}{c}{\textbf{3D Models}} \\
\cmidrule(lr){1-3} \cmidrule(lr){4-6} \cmidrule(lr){7-9}
\textbf{Method} & \textbf{MCQ} & \textbf{Num} &
\textbf{Method} & \textbf{MCQ} & \textbf{Num} &
\textbf{Method} & \textbf{MCQ} & \textbf{Num} \\
\midrule
LLaVA-Next          &  97.40          &  69.06          & MedFlamingo      &  88.35          & \textbf{100.00} & Med3DVLM & \textbf{98.53} & 70.21 \\
GLM-4.1V-9B         & \textcolor{red}{50.25} &  99.99          & LLaVA-Med        &  91.42          & \textbf{100.00} & BTB3D    &  82.54         &  \textbf{82.54}             \\
GLM-4.6V-Flash      &  61.06          & \textcolor{red}{10.91} & HuatuoGPT-Vision & \textcolor{red}{15.97} & \textcolor{red}{0.00} &         M3D      & \textcolor{red}{34.89} &  \textcolor{red}{0.00}         \\
Qwen3-VL 4B         & \textbf{100.00} & \textbf{100.00} & MedMoE           &  47.52          & \textcolor{red}{0.00} &          &                &                \\
Qwen3-VL 8B         & \textbf{100.00} & \textbf{100.00} & Med-VLM R1       &  18.60          & \textcolor{red}{0.00}              &          &                &                \\
InternVL3 8B        & \textbf{100.00} &  94.06          & Med-Gemma 4B     & \textbf{99.85}  & \textbf{100.00} &          &                &                \\
InternVL3 9B        & \textbf{100.00} &  92.73          & UniMedVL         &  99.81          &  97.56          &          &                &                \\
Qwen3.5 9B          &  73.15          &  13.40          & HealthGPT-XL32   &  98.03          & \textbf{100.00} &          &                &                \\
InternVL3.5 14B     & \textbf{100.00} & \textbf{100.00} &                  &                 &                 &          &                &                \\
InternVL3.5 30B-A3B & \textbf{100.00} & \textbf{100.00} &                  &                 &                 &          &                &                \\
GPT 5.4 & \textbf{100.00} & \textbf{100.00} &                  &                 &                 &          &                &                \\
Gemini 2.5 Flash & \textbf{100.00} & \textbf{100.00} &                  &                 &                 &          &                &                \\
\bottomrule
\end{tabular}
\end{table*}

\subsection{Decoding Details}

For benchmarking the multimodal large language models, we set the max new tokens value to 1024, and we disable the sampling method to ensure deterministic outputs, which guarantees reproducibility and enables fair comparison across models.

\subsection{Ablation for slices sampling}
\label{abl:sampling}

The difficulty when adapting 2D vision-language models to volumetric data to explore the spatial understanding is how to project a 3D scan into a fixed set of 2D inputs. From the medical point of view, we conduct experiments on two strategies: (i) \textbf{three orthogonal slices} sampled at the volume center along the axial, coronal, and sagittal axes, the canonical viewing planes used in radiological practice; and (ii) a \textbf{full-slice montage} in which 20 axial slices are tiled into a single grid image to maximize anatomical coverage. The two approaches embody a core trade-off: orthogonal sampling preserves per-slice resolution and aligns with the visual conventions seen during pretraining, whereas the montage exposes the model to a denser sampling of the volume at the cost of substantial per-slice down-scaling.

Table~\ref{tab:ablation_montage} shows that the 3-slice configuration outperforms the full-slices on \emph{average} across all three evaluated models: GPT-5.4~\cite{singh2025openai} (39.23 vs.\ 35.90, +3.33), InternVL3.5-30B~\cite{internvl35} (52.73 vs.\ 51.37, +1.36), and HealthGPT-XL32~\cite{lin2025healthgptmedicallargevisionlanguage} (62.44 vs.\ 41.69, +20.75). The gap is largest for the medical-specialized HealthGPT-XL32, whose pretraining is dominated by single-view 2D radiographs and standard CT/MR projections; tiling perturbs the visual statistics it relies on most. The advantage is also consistent on \textbf{Volume Magnitude} (MRA), the most resolution-sensitive sub-task, where every model scores higher under 3-slice input (e.g., HealthGPT-XL32: 31.11 vs.\ 24.50). The full-slices configuration does occasionally win on individual sub-tasks, GPT-5.4 on DIR (57.14 vs.\ 42.48) and InternVL3.5-30B on EXT and VOL. These results suggest that broader anatomical coverage can aid directional and extent reasoning when resolution loss is tolerable. We therefore adopt the 3-slice orthogonal sampling as the default visual input in all subsequent experiments. However, these gains do not generalize across categories or models: no model achieves a higher average score under the montage as the 2D slices can not provide all of the spatial information to the Language Model; for this reason, we aim to extend the 3D models in the future for further spatial reasoning.

\begin{table*}[t]
\centering
\caption{\textbf{Ablation: 3-Slice vs.\ full slices input.} Comparison of spatial reasoning performance when using three orthogonal slices (axial, coronal, sagittal) versus full slices (20 axial slices tiled in a grid). Each sub-column reports MCA (\%) for a specific reasoning task. \textbf{DIR} = \emph{Directional}, \textbf{EXT} = \emph{Extent / Size / Shape}, \textbf{VOL} = \emph{Volume Magnitude}, \textbf{COMP} = \emph{Comparative}. \textbf{Volume} reports MRA (\%). The best result per model is \textbf{bold}.}
\label{tab:ablation_montage}
\setlength{\tabcolsep}{2.8pt}
\renewcommand{\arraystretch}{1.15}
\small
\resizebox{\textwidth}{!}{%
\begin{tabular}{ll|cccccc|cccccc}
\toprule
\multirow{2}{*}{\textbf{Method}} &
\multirow{2}{*}{\textbf{Size}} &
\multicolumn{6}{c|}{\textbf{3 Slices}} &
\multicolumn{6}{c}{\textbf{Full Slices}} \\
\cmidrule(lr){3-8}\cmidrule(lr){9-14}
 &
 & \textbf{AVG} & \textbf{DIR} & \textbf{EXT} & \textbf{VOL} & \textbf{COMP} & \textbf{Volume}
 & \textbf{AVG} & \textbf{DIR} & \textbf{EXT} & \textbf{VOL} & \textbf{COMP} & \textbf{Volume} \\
\midrule
GPT-5.4 (2026)~\cite{singh2025openai}
 & --
 & \textbf{39.23} & 42.48 & \textbf{39.36} & \textbf{36.26} & \textbf{40.03} & \textbf{41.06}
 & 35.90 & \textbf{57.14} & 36.70 & 28.41 & 40.00 & 40.79 \\
InternVL3.5-30B (2026)~\cite{internvl35}
 & 31B
 & \textbf{52.73} & \textbf{51.63} & 52.85 & 40.28 & \textbf{61.66} & \textbf{36.99}
 & 51.37 & 44.26 & \textbf{53.57} & \textbf{41.25} & 58.29 & 30.56 \\
HealthGPT-XL32 (2025)~\cite{lin2025healthgptmedicallargevisionlanguage}
 & 32B
 & \textbf{62.44} & \textbf{57.80} & \textbf{69.97} & \textbf{55.38} & \textbf{57.53} & \textbf{31.11}
 & 41.69 & 57.14 & 41.38 & 38.52 & 43.53 & 24.50 \\
\bottomrule
\end{tabular}%
}
\end{table*}

\subsection{Limitations and future works}
\label{appendix:limit}
\textbf{Limitations.} Although SpatialMed provides a comprehensive set of question--answer pairs covering volume, size, relative position, and comparative analysis of anatomies and tumors, several medical aspects remain beyond its current scope. In particular, inter-tumor distance and tumor growth rate are not yet incorporated, as constructing such annotations requires extensive and precise numerical labeling by multiple expert radiologists. Moreover, while our study offers clear insights into the limitations of recent MLLMs for medical spatial reasoning, the relatively small number of available medical MLLMs restricts the spatial dimensions we can systematically evaluate, limiting the generalizability of our findings. Finally, our quantitative tasks should be expanded beyond volume estimation to include additional measurements such as distance and mass estimation, which we plan to address in future work.

\textbf{Future Work.} In future work, we plan to broaden the diversity of MCQ tasks to include distance computation, mass-relation comparison, and topological structure comparison. We also intend to investigate KV-cache extensions at inference time and to develop training recipes that help models better overcome the challenges of spatial understanding in the medical domain.
\newpage
\section*{NeurIPS Paper Checklist}

The checklist is designed to encourage best practices for responsible machine learning research, addressing issues of reproducibility, transparency, research ethics, and societal impact. Do not remove the checklist: {\bf The papers not including the checklist will be desk rejected.} The checklist should follow the references and follow the (optional) supplemental material.  The checklist does NOT count towards the page
limit. 

Please read the checklist guidelines carefully for information on how to answer these questions. For each question in the checklist:
\begin{itemize}
    \item You should answer \answerYes{}, \answerNo{}, or \answerNA{}.
    \item \answerNA{} means either that the question is Not Applicable for that particular paper or the relevant information is Not Available.
    \item Please provide a short (1--2 sentence) justification right after your answer (even for \answerNA). 
\end{itemize}

{\bf The checklist answers are an integral part of your paper submission.} They are visible to the reviewers, area chairs, senior area chairs, and ethics reviewers. You will also be asked to include it (after eventual revisions) with the final version of your paper, and its final version will be published with the paper.

The reviewers of your paper will be asked to use the checklist as one of the factors in their evaluation. While \answerYes{} is generally preferable to \answerNo{}, it is perfectly acceptable to answer \answerNo{} provided a proper justification is given (e.g., error bars are not reported because it would be too computationally expensive'' or ``we were unable to find the license for the dataset we used''). In general, answering \answerNo{} or \answerNA{} is not grounds for rejection. While the questions are phrased in a binary way, we acknowledge that the true answer is often more nuanced, so please just use your best judgment and write a justification to elaborate. All supporting evidence can appear either in the main paper or the supplemental material, provided in appendix. If you answer \answerYes{} to a question, in the justification please point to the section(s) where related material for the question can be found.

IMPORTANT, please:
\begin{itemize}
    \item {\bf Delete this instruction block, but keep the section heading ``NeurIPS Paper Checklist"},
    \item  {\bf Keep the checklist subsection headings, questions/answers and guidelines below.}
    \item {\bf Do not modify the questions and only use the provided macros for your answers}.
\end{itemize}


\begin{enumerate}

\item {\bf Claims}
    \item[] Question: Do the main claims made in the abstract and introduction accurately reflect the paper's contributions and scope?
    \item[] Answer: \answerYes{}
    \item[] Justification: The abstract and introduction clearly state our three contributions: (1) the agentic pipeline for spatial VQA generation, (2) the SpatialMed benchmark with 31,253 QA pairs over 2375 CT scans, and (3) the evaluation of 24 state-of-the-art MLLMs, all of which are substantiated in Sections~3 and 4.

\item {\bf Limitations}
    \item[] Question: Does the paper discuss the limitations of the work performed by the authors?
    \item[] Answer: \answerYes{}
    \item[] Justification: Section~3.2 notes that 2D and 3D models receive different input representations, so cross-modality comparisons should be interpreted with caution. The Impact Statement (Appendix A.5) further discusses risks including residual model errors in clinical settings and over-trust in quantitative outputs. Moreover, Section~\ref{appendix:limit} also represent the current limitations of our work and the future work to improve it.

\item {\bf Theory assumptions and proofs}
    \item[] Question: For each theoretical result, does the paper provide the full set of assumptions and a complete (and correct) proof?
    \item[] Answer: \answerNA{}
    \item[] Justification: The paper introduces a benchmark and an evaluation study; it does not contain theoretical results requiring formal proofs.

    \item {\bf Experimental result reproducibility}
    \item[] Question: Does the paper fully disclose all the information needed to reproduce the main experimental results of the paper to the extent that it affects the main claims and/or conclusions of the paper (regardless of whether the code and data are provided or not)?
    \item[] Answer: \answerYes{}
    \item[] Justification: Section~4.1 specifies the model list, zero-shot evaluation protocol, and greedy decoding. Appendices A.1--A.4 and A.6 provide all data-generation prompts, benchmarking prompts, RAG configuration, deterministic answer-parsing rules, and decoding hyperparameters (max\_new\_tokens=1024, sampling disabled).

\item {\bf Open access to data and code}
    \item[] Question: Does the paper provide open access to the data and code, with sufficient instructions to faithfully reproduce the main experimental results, as described in supplemental material?
    \item[] Answer: Yes 
    \item[] Justification: croissant data path \url{https://huggingface.co/datasets/spatialmed/croissant_spatialmed}

\item {\bf Experimental setting/details}
    \item[] Question: Does the paper specify all the training and test details (e.g., data splits, hyperparameters, how they were chosen, type of optimizer) necessary to understand the results?
    \item[] Answer: \answerYes{}
    \item[] Justification: All evaluations are zero-shot (no training is performed). Section~4.1 specifies the inference protocol, and Appendix A.6 details decoding settings (greedy, max\_new\_tokens=1024). Visual input construction (three orthogonal centroid-aligned slices for 2D models, full volume for 3D models) is described in Section~3.2.

\item {\bf Experiment statistical significance}
    \item[] Question: Does the paper report error bars suitably and correctly defined or other appropriate information about the statistical significance of the experiments?
    \item[] Answer: \answerNo{}
    \item[] Justification: Because we evaluate models with greedy decoding (deterministic outputs) on a fixed benchmark, each evaluation produces a single accuracy/MRA value per model and there is no randomness across runs to estimate error bars from. The benchmark is large (31,253 QA pairs) so reported differences across models reflect benchmark-level performance.

\item {\bf Experiments compute resources}
    \item[] Question: For each experiment, does the paper provide sufficient information on the computer resources (type of compute workers, memory, time of execution) needed to reproduce the experiments?
    \item[] Answer: \answerYes{}
    \item[] Justification: In this work, we use 32 CPU workers, and 4xH100s to run in parallel the benchmark for multiple Multimodal Large Language Model
    
\item {\bf Code of ethics}
    \item[] Question: Does the research conducted in the paper conform, in every respect, with the NeurIPS Code of Ethics \url{https://neurips.cc/public/EthicsGuidelines}?
    \item[] Answer: \answerYes{}
    \item[] Justification: The work uses publicly available, de-identified medical imaging datasets under their original licenses, releases only derived spatial QA annotations, and follows the NeurIPS Code of Ethics. Risks and responsible-use considerations are discussed in the Impact Statement (Appendix A.5).

\item {\bf Broader impacts}
    \item[] Question: Does the paper discuss both potential positive societal impacts and negative societal impacts of the work performed?
    \item[] Answer: \answerYes{}
    \item[] Justification: Appendix A.5 (Impact Statement) discusses positive impacts (clinical decision support, more transparent evaluation of medical MLLMs) as well as risks (over-trust in numeric outputs, residual errors in real clinical settings, privacy/data governance considerations).
    
\item {\bf Safeguards}
    \item[] Question: Does the paper describe safeguards that have been put in place for responsible release of data or models that have a high risk for misuse (e.g., pre-trained language models, image generators, or scraped datasets)?
    \item[] Answer: \answerNA{}
    \item[] Justification: We release a benchmark of spatial QA pairs derived from already-public, de-identified medical segmentation datasets; we do not release pretrained generative models, scraped web data, or other assets with a high risk of misuse.

\item {\bf Licenses for existing assets}
    \item[] Question: Are the creators or original owners of assets (e.g., code, data, models), used in the paper, properly credited and are the license and terms of use explicitly mentioned and properly respected?
    \item[] Answer: \answerYes{}
    \item[] Justification: All source datasets (TotalSegmentator, AMOS, Medical Segmentation Decathlon, KiTS, BraTS) and evaluated models (LLaVA-Next, Qwen3-VL, InternVL3/3.5, GLM-4.1V, MedFlamingo, LLaVA-Med, HuatuoGPT-Vision, MedMoE, Med-VLM~R1, Med-Gemma, UniMedVL, HealthGPT, Med-2E3, M3D, BTB3D, Med3DVLM, GPT-5.4, Gemini-2.5-Flash) are cited, and we follow each source's license and terms of use as noted in Appendix A.5.

\item {\bf New assets}
    \item[] Question: Are new assets introduced in the paper well documented and is the documentation provided alongside the assets?
    \item[] Answer: \answerYes{}
    \item[] Justification: We release the SpatialMed benchmark (31,253 QA pairs over 2375 CT scans) and the agentic data construction pipeline at \url{https://github.com/spatialmed/SpatialMed}. Section~3 documents the task taxonomy, anatomical/tumor coverage, generation tools, and validation protocol; Appendices A.1--A.4 document all prompts and parsing rules.

\item {\bf Crowdsourcing and research with human subjects}
    \item[] Question: For crowdsourcing experiments and research with human subjects, does the paper include the full text of instructions given to participants and screenshots, if applicable, as well as details about compensation (if any)?
    \item[] Answer: \answerNA{}
    \item[] Justification: The paper does not involve crowdsourcing or research with human subjects. Validation was conducted by three board-certified radiologist co-authors using the scoring rubric described in Section~3.2 (0/1/2 with average$>$1 inclusion threshold).

\item {\bf Institutional review board (IRB) approvals or equivalent for research with human subjects}
    \item[] Question: Does the paper describe potential risks incurred by study participants, whether such risks were disclosed to the subjects, and whether Institutional Review Board (IRB) approvals (or an equivalent approval/review based on the requirements of your country or institution) were obtained?
    \item[] Answer: \answerNA{}
    \item[] Justification: No new patient data was collected; we use de-identified, publicly available medical imaging datasets that were released under their original IRB/ethical approvals. The study itself does not enroll human subjects.

\item {\bf Declaration of LLM usage}
    \item[] Question: Does the paper describe the usage of LLMs if it is an important, original, or non-standard component of the core methods in this research? Note that if the LLM is used only for writing, editing, or formatting purposes and does \emph{not} impact the core methodology, scientific rigor, or originality of the research, declaration is not required.
    \item[] Answer: \answerYes{}
    \item[] Justification: LLMs are a core component of our agentic data-construction pipeline. Section~3.2 and Appendix A.1 describe the question-generator agent, the clinical validation agent, and three medical specialist agents (InternLM~2, Qwen-3, Llama-3) used for trivial-question filtering, along with the RAG configuration (Appendix A.3) using Qwen3-8B-Embedding and PubMed.

\end{enumerate}
\end{document}